\documentclass[10pt,twocolumn,letterpaper]{article}

\usepackage{authblk}
\usepackage[pagenumbers]{cvpr} % To force page numbers, e.g. for an arXiv version
\usepackage[accsupp]{axessibility}  % Improves PDF readability for those with disabilities.

\usepackage{amsmath}
\usepackage{amssymb}
\usepackage{bm}
\usepackage{multirow}
\usepackage[ruled,vlined]{algorithm2e}
\usepackage{floatrow}

\usepackage{booktabs}
\usepackage{xspace}
\usepackage[pagebackref,breaklinks,colorlinks]{hyperref}
\usepackage{url}

\usepackage{graphicx}
\usepackage{lipsum}
\usepackage{microtype}
\usepackage{nicefrac}
\usepackage{verbatim}
\usepackage{bbding}

\usepackage[capitalize]{cleveref}
\crefname{section}{Sec.}{Secs.}
\Crefname{section}{Section}{Sections}
\Crefname{table}{Table}{Tables}
\crefname{table}{Tab.}{Tabs.}

\makeatletter
\DeclareRobustCommand\onedot{\futurelet\@let@token\@onedot}
\def\@onedot{\ifx\@let@token.\else.\null\fi\xspace}
\def\eg{\emph{e.g}\onedot} 
\def\ie{\emph{i.e}\onedot}

\def\etal{\emph{et al}\onedot}
\renewcommand{\paragraph}{%
	\@startsection{paragraph}{4}{\z@}%
	{0.1em \@plus 0.5ex \@minus 0.2ex}{-1em}%
	{\normalsize\bf}%
}
\newcommand\rurl[1]{%
  \href{https://#1}{\nolinkurl{#1}}%
}
\makeatother

\newcommand{\bh}{\bm{h}}
\newcommand{\bx}{\bm{x}}
\newcommand{\by}{\bm{y}}

\newcommand{\SL}{{Supervised Learning}}
\newcommand{\SSL}{{Semi-supervised Learning}}
\newcommand{\DM}{{Direct-Map}}
\newcommand{\PM}{{Perceptual-Map}}
\newcommand{\CL}{{Contrastive-Loss}}
\newcommand{\DL}{{Discriminator-Loss}}

\begin{document}

\title{Learning Multiple Dense Prediction Tasks from Partially Annotated Data}

\author[]{\vspace{-0.3cm}Wei-Hong Li}
\author[]{Xialei Liu}
\author[]{Hakan Bilen\vspace{-0.25cm}}

\affil[]{VICO Group, University of Edinburgh, United Kingdom\vspace{-0.25cm}}
\affil[]{\small \rurl{github.com/VICO-UoE/MTPSL}\vspace{-0.3cm}}

\maketitle

%%%%%%%%% ABSTRACT
\begin{abstract}
    Despite the recent advances in multi-task learning of dense prediction problems, most methods rely on expensive labelled datasets.
    In this paper, we present a label efficient approach and look at jointly learning of multiple dense prediction tasks on partially annotated data (\ie not all the task labels are available for each image), which we call {multi-task partially-supervised learning}. 
    We propose a multi-task training procedure that successfully leverages task relations to supervise its multi-task learning when data is partially annotated. 
    In particular, we learn to map each task pair to a {joint pairwise task-space} which enables sharing information between them in a computationally efficient way through another network conditioned on task pairs, and avoids learning trivial cross-task relations by retaining high-level information about the input image.
    We rigorously demonstrate that our proposed method effectively exploits the images with unlabelled tasks and outperforms existing semi-supervised learning approaches and related methods on three standard benchmarks.

\end{abstract}

%%%%%%%%% BODY TEXT
\section{Introduction}\label{sec:intro}
% !TEX root = main.tex

With the recent advances in dense prediction computer vision problems~\cite{long2015fully,he2017mask,eigen2014depth,zhang2020select,zhang2020uc,xia2017joint,yu2017casenet,su2021pixel,minaee2021image,wang2021end,tian2020conditional,poggi2020uncertainty}, where the aim is to produce pixel-level predictions (\eg semantic and instance segmentation, depth estimation), the interest of the community has started to shift towards the more ambitious goal of learning multiple of these problems jointly by multi-task learning (MTL)~\cite{caruana1997multitask}.
Compared to the standard single task learning (STL) that focuses on learning an individual model for each task, MTL aims at learning a single model for multiple tasks with a better efficiency and generalization tradeoff while sharing information and computational resources across them.

\begin{figure}[h!]
\begin{center}
\includegraphics[width=0.9\linewidth]{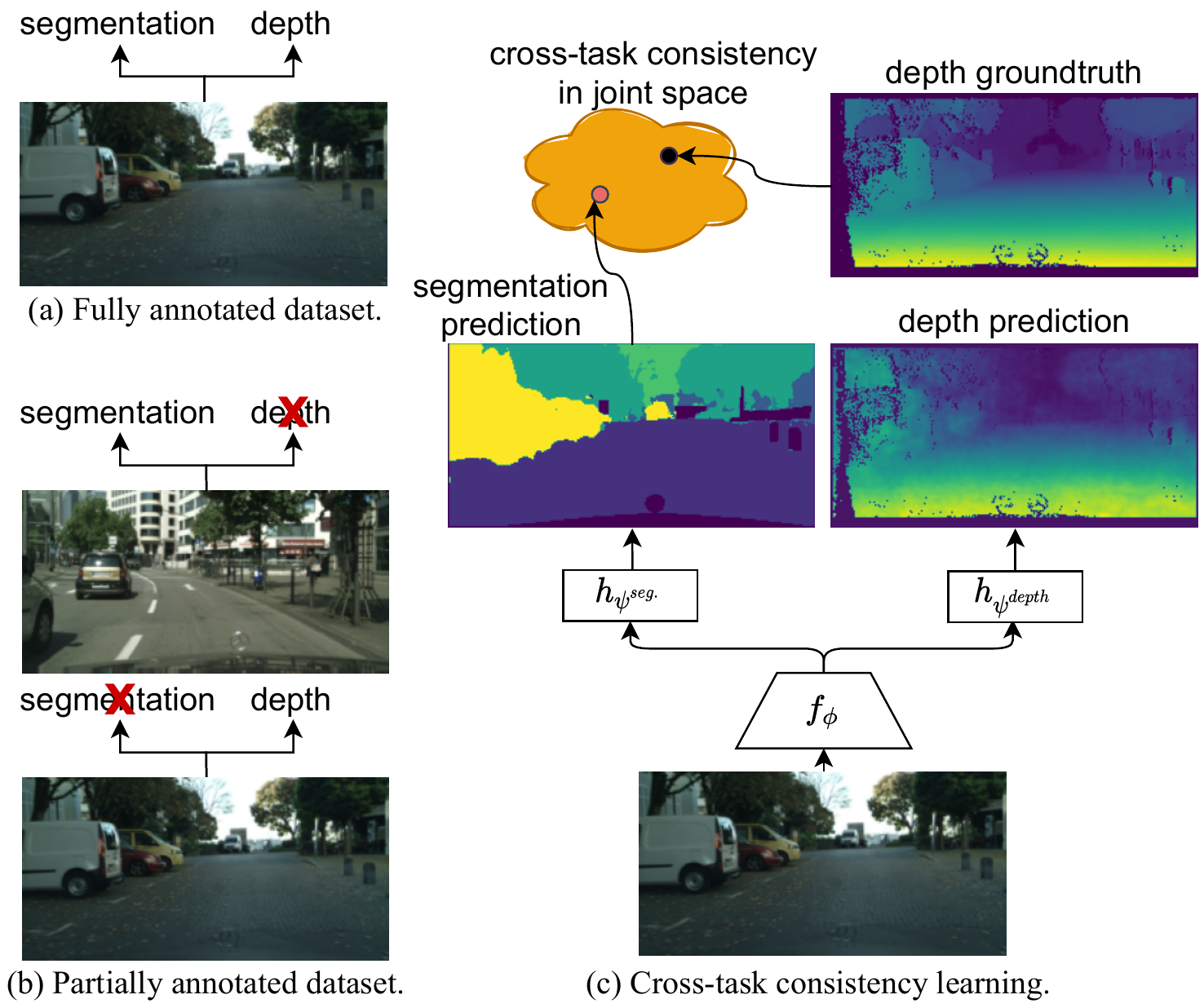}
\end{center}
\vspace{-0.65cm}
\caption{\textbf{Multi-task partially supervised learning.} We look at the problem of learning multiple tasks from partially annotated data (b) where not all the task labels are available for each image, which generalizes over the standard supervised learning (a) where all task labels are available.
We propose a MTL method that employs a shared feature extractor ($f_\phi$) with task-specific heads ($h_{\psi}$) and exploits label correlations between each task pair by mapping them into a \emph{joint pairwise task-space} and penalizing inconsistencies between the provided ground-truth labels and predictions (c).}
\label{fig:mtssl}
\end{figure}

Recent MTL dense prediction methods broadly focus on designing MTL architectures~\cite{misra2016cross,ruder2019latent,gao2019nddr,liu2019end,lu2017fully,vandenhende2019branched,bruggemann2020automated,guo2020learning,bragman2019stochastic,xu2018pad,zhang2019pattern,zhang2018joint,vandenhende2020mti,zhou2020pattern} that enable effective sharing of information across tasks and improving the MTL optimization~\cite{gong2019comparison,kendall2018multi,chen2018gradnorm,liu2019end,guo2018dynamic,sener2018multi,lin2019pareto,yu2020gradient,chen2020just,li2020knowledge} to balance the influence of each task-specific loss function and to prevent interference between the tasks in training.
We refer to \cite{vandenhende2021multi} for a more comprehensive review.
One common and strong assumption in these works is that each training image has to be labelled for all the tasks (\cref{fig:mtssl}(a)).
There are two main practical limitations to this assumption.
First, curating multi-task image datasets (\eg KITTI~\cite{geiger2012we} and CityScapes~\cite{cordts2016cityscapes}) typically involves using multiple sensors to produce ground-truth labels for several tasks, and obtaining all the labels for each image requires very accurate synchronization between the sensors, which is a challenging research problem by itself \cite{voges2018timestamp}.
Second, imagine a scenario where one would like to add a new task to an existing image dataset which is already annotated for another task and obtaining the ground-truth labels for the new task requires using a different sensor (\eg depth camera) to the one which is used to capture the original data.
In this case, labelling the previously recorded images for the new task will not be possible for many visual scenes (\eg uncontrolled outdoor environments).
Such real-world scenarios lead to obtaining partially annotated data and thus ask for algorithms that can learn from such data.

In this paper, we look at a more realistic and general case of the MTL dense prediction problem where not all the task labels are available for each image (\cref{fig:mtssl}(b)) and we call this setting \emph{multi-task partially supervised learning}.
In particular, we assume that each image is at least labelled for one task and each task at least has few labelled images
% there are at least few images labelled for each task 
and we would like to learn a multi-task model on them.
A naive way of learning from such partial supervision is to train a multi-task model only on the available labels (\ie by setting the weight of the corresponding loss function to 0 for the missing task labels).
Though, in this setting, the MTL model is trained on all the images thanks to the parameter sharing across the tasks, it cannot extract the task-specific information from the images for the unlabelled tasks.
To this end, one can extend existing single-task semi-supervised learning methods to MTL by penalizing the inconsistent predictions of images over multiple perturbations for the unlabelled tasks (\eg \cite{chen2020multi,liu2008semi,terzopoulos2019semi,latif2019multi,imran2020partly}).
While this strategy ensures consistent predictions over various perturbations, it does not guarantee consistency across the related tasks. 

An orthogonal information that has recently been used in MTL is cross-task relation \cite{zamir2020robust,lu2021taskology,saha2021learning} which aims at producing consistent predictions across task pairs.
Unfortunately existing methods are not directly applicable for learning from partial supervision, as they require either each training image to be labelled with all the task labels \cite{zamir2020robust,saha2021learning} or cross-task relations that can be analytically derived~\cite{lu2021taskology}.
In our setting, compared to \cite{zamir2020robust,lu2021taskology,saha2021learning}, there are fewer training images available with ground-truth labels of each task pair and thus it is harder to learn the relationship.
In addition, unlike \cite{lu2021taskology}, we focus on the general setting where one task label cannot be accurately obtained from another (\eg from semantic segmentation to depth) and hence learning exact mappings between two task labels is not possible.

Motivated by these challenges, we propose a MTL approach that shares a feature extractor between tasks and also learns to relate each task pair in a learned \emph{joint pairwise task-space} (illustrated in~\cref{fig:mtssl}(c)), which encodes only the shared information between them and does not require the ill-posed problem of recovering labels of one task from another one.
There are two challenges to this goal.
First, a naive learning of the joint pairwise task-spaces can lead to trivial mappings that take all predictions to the same point such that each task produces artificially consistent encodings with each other.
To this end, we regulate learning of each mapping by penalizing its output to retain high-level information about the input image.
Second, the computational cost of modelling each task pair relation can get exponentially expensive with the number of tasks.
To address this challenge, we use a single encoder network to learn all the pairwise-task mappings, however, dynamically estimate its weights by conditioning them on the target task pair.

The main contributions of our method are as following. 
We propose a new and practical setting for multi-task dense prediction problems and a novel MTL model that penalizes cross-task consistencies between pairs of tasks in joint pairwise task-spaces, each encoding the commonalities between pairs, in a computationally efficient manner.
We show that our method can be incorporated to several architectures and significantly outperforms the related baselines in three standard multi-task benchmarks.

\section{Related Work}\label{sec:rel}
% !TEX root = main.tex

\paragraph{Multi-task Supervised Learning.}
Multi-task Learning (MTL)~\cite{caruana1997multitask,vandenhende2021multi,ruder2017overview,zhang2017survey} aims at learning a single model that can infer all desired task outputs given an input. 
The prior works can be broadly divided into two groups.
The first one~\cite{misra2016cross,ruder2019latent,gao2019nddr,liu2019end,lu2017fully,vandenhende2019branched,bruggemann2020automated,guo2020learning,bragman2019stochastic,xu2018pad,zhang2019pattern,zhang2018joint,vandenhende2020mti,zhou2020pattern,bruggemann2021exploring} focuses on improving network architecture by better sharing information across tasks and learning task-specific representation by devising cross-task attention mechanism~\cite{misra2016cross}, task-specific attention modules~\cite{liu2019end}, gating strategies~\cite{bruggemann2020automated,guo2020learning}, etc.
The second one aims to improve the imbalanced optimization problem caused by jointly optimizing different losses of various tasks as the difficulty levels, loss magnitudes, and characteristics of tasks are various. 
To this end, the recent work~\cite{gong2019comparison,kendall2018multi,chen2018gradnorm,liu2019end,guo2018dynamic,sener2018multi,lin2019pareto,yu2020gradient,chen2020just} enable a more balanced optimization for multi-task learning network by dynamically adjusting weights of each loss functions based on task-certainty~\cite{kendall2018multi}, Pareto optimality~\cite{sener2018multi}, discarding conflicting gradient components~\cite{yu2020gradient}, etc.
However, these works focus on the supervised setting, where each sample in the dataset is annotated for all desired tasks.

\paragraph{Multi-task Semi-supervised Learning.}
Learning multi-task model on fully annotated data would require large-scale labeled data and it is costly to collect sufficient labeled data. 
Thus few works propose to learn multi-task learning model using semi-supervised learning strategy~\cite{liu2008semi,zhang2009semi,wang2009semi,chen2020multi,liu2008semi,terzopoulos2019semi,latif2019multi,imran2020partly} and they assume that the dataset consists of limited data annotated with all tasks labels and a large amount of unlabeled data. 
Liu~\etal~\cite{liu2008semi} extend single-task semi-supervised learning to multi-task learning by learning a classifier per task jointly under the constraint of a soft-sharing prior imposed over the parameters of the classifiers. 
In \cite{chen2020multi,liu2008semi,terzopoulos2019semi,latif2019multi,imran2020partly}, the authors employ a regularization term on the unlabeled samples of each tasks that encourages the model to produce `consistent' predictions when its inputs are perturbed.

\paragraph{Cross-task Relations.}
A rich body of work~\cite{liu2010single,bilen2016integrated,zamir2016generic,zamir2018taskonomy,lu2021taskology,zamir2020robust,saha2021learning,zhou2017unsupervised,casser2019depth,wang2021domain,hoyer2021improving,sun2021see,hoiem2008closing,martin2004learning} study the relations between tasks in MTL. 
Most related to ours, \cite{saha2021learning} explore the relations between segmentation and depth and propose a better fusion strategy to fuse two tasks predictions for domain adaptation. 
Zamir~\etal~\cite{zamir2020robust} study the cross-task consistency learned from groundtruth of all tasks for robust learning, \ie the predictions made for multiple tasks from the same image are not independent, and therefore, are expected to be `consistent'. 
Similar to~\cite{zamir2020robust}, Lu \etal~\cite{lu2021taskology} propose to leverage the cross-task consistency between predictions of different tasks on unlabeled data in a mediator dataset when jointly learning multiple models for distributed training. 
To regularize the cross-task consistency, Lu~\etal~\cite{lu2021taskology} design consistency losses according to the consistency between adjacent frames in videos, relations between depth and surface normal, etc. 
In this paper, we also exploit the cross-task consistency in MTL, however, from partially annotated data where the mapping from one task label to another cannot be analytically derived or exactly learned.
To this end, unlike \cite{zamir2020robust,lu2021taskology}, we learn a joint task-space for each task pair rather than measuring consistency in one's task space.
Finally, our method learns cross-task in a more computationally efficient way than \cite{zamir2020robust,lu2021taskology} by sharing parameters across different mappings and conditioning its output on the related task-pair.

\section{Method}\label{sec:method}
% !TEX root = main.tex

\begin{figure*}[t]
\begin{center}
\includegraphics[width=0.92\linewidth]{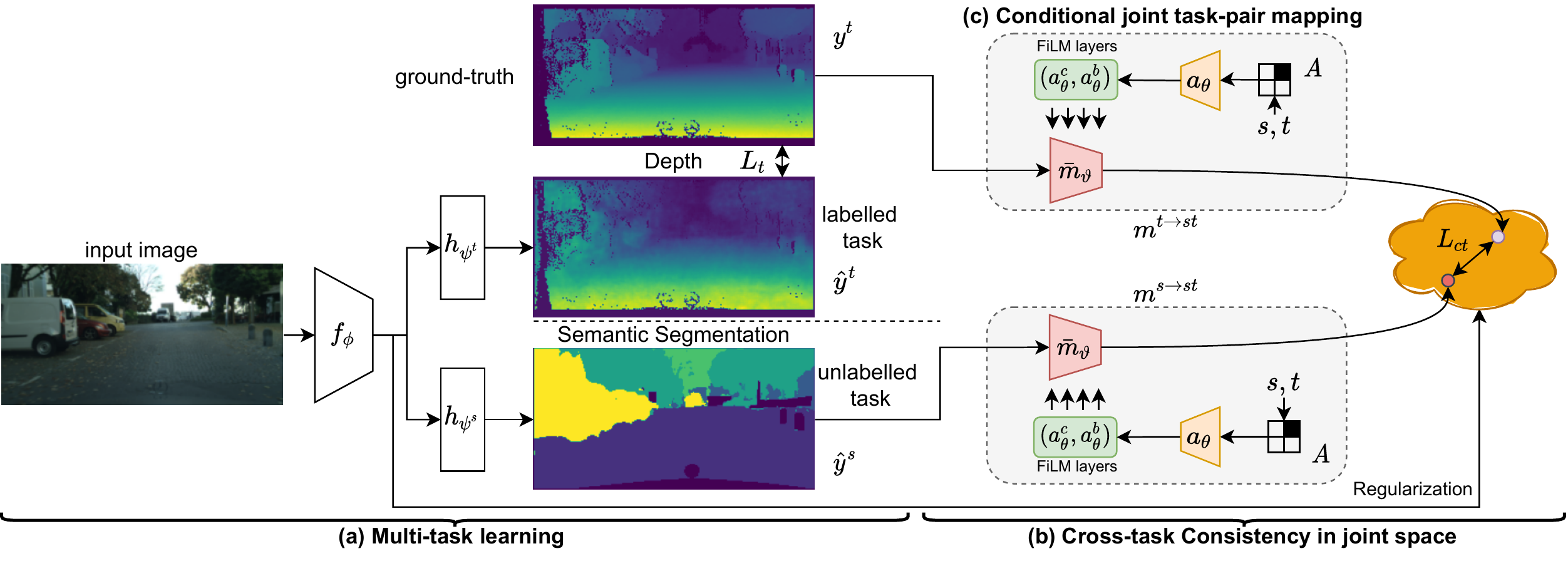}
\end{center}
\vspace{-0.75cm}
\caption{Illustration of our method for multi-task partially supervised learning. Given an image, our method uses a shared feature extractor $f_{\phi}$ taking in the input image and task-specific decoders ($h_{\psi^{s}}$ and $h_{\psi^{t}}$) to produce predictions for all tasks (a). We compute the supervised loss $L_{t}$ for labelled task. Besides, we regularize the cross-task consistency $L_{ct}$ between the unlabelled task's prediction $\hat{\by}^s$ and the labelled task's ground-truth $\by^t$ in a joint space for the unlabelled task (b). To learn the cross-task consistency efficiently, we propose to use a shared mapping function whose output is conditioned on the task-pair (c) and regularize the learning of mapping function using the feature from $f_{\phi}$ to prevent trivial solution.
}
\label{fig:framework}
\end{figure*}

\subsection{Problem setting}
% definition
Let $\bx \in \mathbb{R}^{3 \times H\times W}$ and $\by^{t} \in \mathbb{R}^{O^{t} \times H \times W} $ denote an $H\times W$ dimensional RGB image and its dense label for task $t$ respectively, where $O^{t}$ is the number of output channels for task $t$.
Our goal is to learn a function $\hat{y}^{t}$ for each task $t$ that accurately predicts the ground-truth label $\by^{t}$ of previously unseen images.
While such a task-specific function can be learned for each task independently, a more efficient design is to share most of the computations across the tasks via a common feature encoder, convolutional neural network $f_{\phi}: \mathbb{R}^{3 \times H\times W} \rightarrow \mathbb{R}^{C \times H'\times W'}$ parameterized by $\phi$ that takes in an image and produces a $C$ feature maps, each with $H'\times W'$ resolution, where typically $H'<H$ and $W'<W$.
In this setting, $f_{\phi}$ is followed by multiple task-specific decoders $h_{\psi^{t}}: \mathbb{R}^{C \times H'\times W'} \rightarrow \mathbb{R}^{O^{t} \times H \times W}$, each with its own task-specific weights $\psi^{t}$ that decodes the extracted feature to predict the label for the task $t$, \ie $\hat{y}^{t}(\bx)=h_{\psi^{t}}\circ f_{\phi}(\bx)$ (\cref{fig:framework}(a)).

% more definition
Let $\mathcal{D}$ denote a set of $N$ training images with their corresponding labels for $K$ tasks.
Assume that for each training image $\bx$, we have ground-truth labels available only for some tasks where we use $\mathcal{T}$ and $\mathcal{U}$ to store the indices of labeled and unlabelled tasks respectively, where $|\mathcal{T}|+|\mathcal{U}|=K$, $\mathcal{U}=\varnothing$ indicates all labels available for $\bx$ and $\mathcal{T}=\varnothing$ indicates no labels available for $\bx$.
In this paper, we focus on the partially annotated setting, where each image is labelled at least for one task ($|\mathcal{T}|\ge 1$) and each task at least has few labelled images.

% multi-task supervised learning
A naive way of learning $\hat{y}^{t}$ for each task on the partially annotated data $\mathcal{D}$ is to jointly optimize its parameters on the labelled tasks as following:
\begin{equation}\label{eq:mtl}
    \min_{\phi, \psi}  \frac{1}{N}\sum_{n=1}^{N}\frac{1}{|\mathcal{T}_{n}|}\sum_{t \in \mathcal{T}_{n}}L^{t}(\hat{y}^{t}(\bx_{n}), \by^{t}_{n}),
\end{equation} where ${n}$ is the image index and $L^{t}$ is the task-specific differentiable loss function. 
We denote this setting as the (vanilla) MTL.
Here, thanks to the parameter sharing through the feature extractor, its task-agnostic weights are learned on all the images.
However, the task-specific weights $\psi^{t}$ are trained only on the labeled images.

A common strategy to exploit such information from unlabeled tasks is to formulate the problem in a semi-supervised learning (SSL) setting.
Recent successful SSL techniques~\cite{berthelot2019mixmatch,sohn2020fixmatch} focus on learning models that can produce consistent predictions for unlabelled images when its input is perturbed in various ways.
\begin{equation}\label{eq:mtlsemi}
\begin{aligned}
    \min_{\phi, \psi} & \frac{1}{N}\sum_{n=1}^{N}\Big(\frac{1}{|\mathcal{T}_{n}|}\sum_{t \in \mathcal{T}_{n}}L^{t}(\hat{y}^{t}(\bx_{n}), \by_{n}^{t}) \\
    &+\frac{1}{|\mathcal{U}_{n}|}\sum_{t\in \mathcal{U}}L_{u}(e_r(\hat{y}^{t}(\bx_{n})),\hat{y}^{t}(e_r(\bx_{n}))\Big),
\end{aligned}
\end{equation} where $L_{u}$ is the unsupervised loss function and $e_r$ is a geometric transformation (\ie cropping) parameterized by the random variable $r$ (\ie bounding box location).
In words, for the unsupervised part, we apply our model to the original input $\bx$ and also its cropped version $e_r(\bx)$, and then we also crop the prediction corresponding to the original input $e_r(\hat{y}^{t}(\bx_{n}))$ before we measure the difference between two by using $L_{u}$.
Note that we are aware of more sophisticated task-specific SSL methods for semantic segmentation~\cite{olsson2021classmix,mendel2020semi}, depth estimation~\cite{kuznietsov2017semi,guizilini2020robust}, however, combining them for multiple tasks, each with different network designs and learning formulations is not trivial and here we focus on one SSL strategy that uses one perturbation type (\ie random cropping) and $L_{u}$ (\ie  mean square error) can be applied to several tasks.

\subsection{Cross-task consistency learning}
% cross-task consistency learning
While optimizing \cref{eq:mtlsemi} allows learning both task-agnostic and task-specific weights on the labeled and unlabelled data, it does not leverage cross-task relations, which can be used to further supervise unlabelled tasks.
Prior works~\cite{zamir2020robust,lu2021taskology} define the cross-task relations by a mapping function $m^{s \rightarrow t}$ for each task-pair $(s,t)$ which maps the prediction for the source task $s$ to target task $t$ labels.
The mapping function in \cite{lu2021taskology} is analytical based on the assumption that target task labels can be analytically computed from source labels.
While such analytical relations is possible only for certain task pairs, each mapping function in \cite{zamir2020robust} is parameterized by a deep network and its weights are learned by minimizing $L_{ct}( m^{s \rightarrow t}(\by^s),\by^t$), where $L_{ct}$ is cross-task function that measures the distance between the mapped source labels and target labels.
There are two limitations to this method in our setting.
First the training set has limited labelled number of images for both source and target tasks ($\by^s$ and $\by^t$).
Second learning such pairwise mappings accurately is not often possible in our case, as the labels of one task can only be partially recovered from another task (\eg semantic segmentation to depth estimation).
Note that this ill-posed problem can be solved accurately when strong prior knowledge about the data is available.

To employ cross-task consistency to our setting, we map each task pair $(s,t)$ to a lower-dimensional joint pairwise task-space where only the common features of both tasks are encoded (\cref{fig:framework}(b)).
Formally, each pairwise task-space for $(s,t)$ is defined by a pair of mapping functions, $m_{\vartheta_{s}^{st}}: \mathbb{R}^{O^s \times H \times W} \rightarrow \mathbb{R}^D$ and $m_{\vartheta_{t}^{st}}: \mathbb{R}^{O^t \times H \times W} \rightarrow \mathbb{R}^D$ parameterized by ${\vartheta_{s}^{st}}$ and ${\vartheta_{t}^{st}}$ respectively.
The cross-task consistency can be incorporated to \cref{eq:mtl} as following:
\begin{equation}\label{eq:mtlct}
    \begin{aligned}
        \min_{\phi, \psi, \vartheta} & \frac{1}{N}\sum_{n=1}^{N}\Big(\frac{1}{|\mathcal{T}_{n}|}\sum_{t \in \mathcal{T}_{n}}L^{t}\big(\hat{y}^{t}(\bx_{n}), \by^{t}_{n}\big) + \\
        &\frac{1}{|\mathcal{U}_{n}|}\sum_{s\in \mathcal{U}_{n}, t \in \mathcal{T}_{n}}L_{ct}\big(m_{\vartheta_{s}^{st}}(\hat{y}^{s}(\bx_{n})),  m_{\vartheta_{t}^{st}}(\by_{n}^t)\big)\Big),
    \end{aligned}
\end{equation} where $L_{ct}$ is cosine distance (\ie $L_{ct}(\mathbf{a},\mathbf{b})=1-(\mathbf{a} \cdot \mathbf{b})/(|\mathbf{a}||\mathbf{b}|$).
In words, along with the MTL optimization, \cref{eq:mtlct} minimizes the cosine distance between the embeddings of the unlabelled task prediction $\hat{\by}_{s}$ and the annotated task label $\by^t$ in the joint pairwise task space.
Here $m_{\vartheta_{s}^{st}}$ and $m_{\vartheta_{t}^{st}}$ are not necessarily equal to allow for treating the mapping from predicted and ground-truth labels differently.
Note that one can also include the semi-supervised term $L_{u}$ in \cref{eq:mtlct}.
However we empirically found that it does not bring any tangible performance gain when used with the cross-task term $L_{ct}$.

There are two challenges to learn non-trivial pairwise mapping functions in a computationally efficient way.
First the number of pairwise mappings to learn quadratically grows with the number of tasks.
Although the mapping functions are only used in training, it can still be computationally expensive to train many of them jointly.
In addition, learning an accurate mapping for each task-pair can be challenging in case of limited labels.
Second the mapping functions can simply learn a trivial solution such that each task is mapped to a fixed point (\eg zero vector) in the joint space.

\paragraph{Conditional joint task-pair mapping.}
To address the first challenge, as shown in \cref{fig:framework}(c), we propose to use a task-agnostic mapping function $\bar{m}_{\vartheta}$ with one set of parameters $\vartheta$ whose output is conditioned both on the input task ($s$ or $t$) and task-pair ($s,t$) through an auxiliary network ($a_\theta$).
Concretely, let $A$ denote a variable that includes the input task ($s$ or $t$) and target pair $(s,t)$ for a pairwise mapping which in practice we encode with an asymmetric $K\times K$ dimensional matrix by setting the corresponding entry to 1 (\ie $A[s,t]=1$ or $A[t,s]=1$) and the other entries to 0.
Note that the diagonal entries are always zero, as we do not define any self-task relation.
Let $\bar{m}_{\vartheta}$ be a multi-layer network and $\bh_i$ denote a $M$ channel feature map of its $i$-th layer for which the auxiliary network $a_{\theta}$, parameterized by $\theta$, takes in $A$ and outputs two $M$-dimensional vectors $a^c_{\theta,i}$ and $a^b_{\theta,i}$.
These vectors are applied to transform the feature map $\bh_i$ in a similar way to ~\cite{perez2018film} as following:
\[
    {\bh}_i \leftarrow a^c_{\theta,i}(A) \odot  \bh_i + a^b_{\theta,i}(A)
\]
where $\odot$ denote a Hadamard product.
In words, the auxiliary network alters the output of the task-agnostic mapping function $\bar{m}_{\vartheta}$ based on $A$.
For brevity, we denote the conditional mapping from $s$ to $(s,t)$ as $m^{s\rightarrow st}$ which is a function of $\bar{m}_{\vartheta}$ and $a_{\theta}$ and hence parameterized with $\vartheta$ and $\theta$.

We implement each $a_i^c$ and $a_i^b$ as an one layer fully-connected network.
Hence, given the light-weight auxiliary network, the computational load for computing the conditional mapping function, in practice, does not vary with the number of task-pairs.
Finally, as the dimensionality of each task label vary -- \eg while $O^t$ is 1 for depth estimation and $O^t$ equals to number of categories in semantic segmentation --, we use task-specific input layers and pass each prediction to the corresponding one before feeding it to the joint pairwise task mapping.
In the formulation, we include these layers in our mapping $\bar{m}_{\vartheta}$ and explain their implementation details in \cref{sec:exp}.

\paragraph{Regularizing mapping function.} 
To avoid learning trivial mappings, we propose a regularization strategy (\cref{fig:framework}) that encourages the mapping to retain high-level information about the input image.
To this end, we penalize the distance between the output of the mapping function and a feature vector that is extracted from the input image.
In particular, we use the output of the task-agnostic feature extractor $f_\phi(\bx)$ in the regularization.
Now we can add the regularizer to the formulation in \cref{eq:mtlct}:
\begin{equation}\label{eq:mtlours}
\begin{aligned}
    \min_{\phi, \psi, \vartheta, \theta} & \frac{1}{N}\sum_{n=1}^{N}
    \Big(
        \frac{1}{|\mathcal{T}_{n}|}\sum_{t \in \mathcal{T}_{n}}L^{t}\big(\hat{y}^{t}(\bx_{n}), \by^{t}_{n}\big) + \\
    &\frac{1}{|\mathcal{U}_{n}|} \sum_{s\in \mathcal{U}_{n}, t \in \mathcal{T}_{n}}L_{ct}
    \big(
        m^{s\rightarrow st}(\hat{y}^{s}(\bx_{n})), m^{t\rightarrow st}(\by_{n}^t)\big)
     \\
    & + R(f_{\phi}(\bx_{n}), m^{s\rightarrow st}(\hat{y}_s(\bx_{n}))) \\
    & + R(f_{\phi}(\bx_{n}), m^{t\rightarrow st}(\by^{t}_{n}))
    \Big), \\
\end{aligned}
\end{equation} where $f_{\phi}(\bx)$ is the feature from feature encoder $f_{\phi}$, $R$ is the loss function and we use the cosine similarity loss for $R$ in this work. 

\paragraph{Alternative mapping strategies.}
Here we discuss two different mapping strategies to exploit cross-task consistency proposed in \cite{zamir2020robust} and their adoption to our setting.
As both require learning a mapping from one task's groundtruth label to another one and we have either no or few images with both groundtruth labels, here we approximate them by learning mappings from prediction of one task to another task's groundtruth.
In the first case, one can substitute our cross-consistency loss and regularization terms with $L_{ct}( m^{s \rightarrow t}(\hat{y}^s(\bx)),\by^t)$ in \cref{eq:mtlours}, which is denoted as \emph{\DM}.
In the second case, we replace our terms with $L_{ct}( m^{s \rightarrow t}(\hat{y}^s(\bx)),m^{s \rightarrow t}(\by^s))$ that maps both the groundtruth $\by^s$ and predicted labels $\hat{\by}^s$ and minimize their distance in task $t$'s label space.
We denote this setting as \emph{\PM} and compare to them in \cref{sec:exp}.

\paragraph{Alternative loss and regularization strategies.}
Alternatively, our cross-consistency loss and regularization terms can be replaced with another loss function only that does not allow for learning of trivial mappings.
One such loss function is contrastive loss where one can define the predictions for two tasks on the same image as a positive pair (\ie $m^{s\rightarrow st}(\hat{y}^s(\bx_i))$ and $m^{t\rightarrow st}(\by^{t}_i)$) and on different images as a negative pair (\ie $m^{s\rightarrow st}(\hat{y}^s(\bx_j))$ and $m^{t\rightarrow st}(\by^{t}_i)$), and penalize when the distance from the positive one is bigger than the negative one.
We denote this setting as \emph{\CL}.
Another method which also employs positive and negative pairs involves using a discriminator network. 
The discriminator (a convolutional neural network) takes in positive and negative pairs and predicts their binary labels, while the parameters of the MTL network and mapping functions are alternatively optimized.
We denote this setting as \emph{\DL} and compare to the alternative methods in \cref{sec:exp}.

\section{Experiments}\label{sec:exp}
% !TEX root = main.tex

\paragraph{Datasets.}
We evaluate all methods on three standard dense prediction benchmarks, Cityscapes~\cite{cordts2016cityscapes}, NYU-V2~\cite{silberman2012indoor}, and PASCAL~\cite{everingham2010pascal}.
Cityscapes~\cite{cordts2016cityscapes} consists of street-view images, which are labeled for two tasks: 7-class semantic segmentation\footnote{The original version of Cityscapes provides labels 19-class semantic segmentation. We follow the evaluation protocol in \cite{liu2019end}, we use labels of 7-class semantic segmentation. Please refer to \cite{liu2019end} for more details.} and depth estimation. We resize the images to $128 \times 256$ to speed up the training as~\cite{liu2019end}.
NYU-V2~\cite{silberman2012indoor} contains RGB-D indoor scene images, where we evaluate performances on 3 tasks, including 13-class semantic segmentation, depth estimation, and surface normals estimation. We use the true depth data recorded by the Microsoft Kinect and surface normals provided in \cite{eigen2015predicting} for depth estimation and surface normal estimation. All images are resized to $288 \times 384$ resolution as in \cite{liu2019end}.
PASCAL~\cite{everingham2010pascal} is a commonly used benchmark for dense prediction tasks. We use the data splits from PASCAL-Context~\cite{chen2014detect} which has annotations for semantic segmentation, human part segmentation and semantic edge detection. Additionally, as in~\cite{vandenhende2021multi}, we also consider the tasks of surface normals prediction and saliency detection and use the annotations provided by \cite{vandenhende2021multi}.

\paragraph{Experimental setting.}
For the evaluation of multi-task models learned in different partial label regimes, we design two settings: \textit{(i)} \emph{random} setting where,  we randomly select and keep labels for at least 1 and at most $K-1$ tasks where $K$ is the number of tasks, \textit{(ii)} \emph{one} label setting, where we randomly select and keep label only for 1 task for each training image.

In Cityscapes and NYU-v2, we follow the training and evaluation protocol in~\cite{liu2019end} and we use the the SegNet~\cite{badrinarayanan2017segnet} as the MTL backbone for all methods. 
As in ~\cite{liu2019end}, we use cross-entropy loss for semantic segmentation, l1-norm loss for depth estimation in Cityscapes, and cosine similarity loss for surface normal estimation in NYU-v2. 
We use the exactly same hyper-parameters including learning rate, optimizer and also the same evaluation metrics,  mean intersection over union (mIoU), absolute error (aErr) and mean error (mErr) in the predicted angles to evaluate the semantic segmentation, depth estimation and surface normals estimation task, respectively in~\cite{liu2019end}. 
We use the encoder of SegNet for the joint pairwise task mapping ($\bar{m}_{\vartheta}$) and one convolutional layer as task specific input layer in $\bar{m}_{\vartheta}$. 
For \texttt{Direct-Map} and \texttt{Perceptual-Map}, as in~\cite{zamir2020robust} we use the whole SegNet as the cross-task mapping functions.

In PASCAL, we follow the training, evaluation protocol and implementation in~\cite{vandenhende2021multi} and employ the ResNet-18~\cite{he2016deep} as the encoder shared across all tasks and Atrous Spatial Pyramid Pooling (ASPP)~\cite{chen2018encoder} module as task-specific heads. 
We use the same hyper-parameters, \eg learning rate, augmentation, loss functions, loss weights in \cite{vandenhende2021multi}.
For evaluation metrics, we use the optimal dataset F-measure (odsF)~\cite{martin2004learning} for edge detection, the standard mean intersection over union (mIoU) for semantic segmentation, human part segmentation and saliency estimation are evaluated,  mean error (mErr) for surface normals.
We modify the ResNet-18 to have task specific input layers (one convolutional layer for each task) before the residual blocks as the mapping function $\bar{m}_{\vartheta}$ in our method.
We refer to the supplementary for more details. 
% Note that our code and models will be made public based upon acceptance.

\subsection{Results}
We compare our method to multiple baselines including the vanilla MTL \SL~(SL) baseline in \cref{eq:mtl} on both all the labels and partial labels in \cref{eq:mtl}, and the MTL \SSL~(SSL) in \cref{eq:mtlsemi}, also variations of our method with \DM, \PM, \CL~and \DL~as described in \cref{sec:method}. 
We use uniform weights for task-specific losses for all, unless stated otherwise.

\paragraph{Results on Cityscapes.}
We first compare our method to the baselines on Cityscapes in \cref{tab:citys} for only \emph{one} label setting as there are two tasks in total. 
The results of MTL model learned with SL when all task labels are available for training to serve as a strong baseline.
In the partial label setting (one task label per image), the performance of the SL baseline drops substantially compared to its performance in full supervision setting. 
While the SSL baseline, by extracting task-specific information from unlabelled tasks, improves over SL, further improvements are obtained by exploiting cross-task consistency in various ways except \DL.
The methods learn mappings from one task to another one (\PM\ and \DM) surprisingly perform better than the ones learning joint space mapping functions (\CL\ and \DL), possibly due to insufficient number of negative samples.
Due to the same reason, we exclude the further comparisons to \CL\ and \DL\ in NYU-v2 and PASCAL and include them in the supplementary.
Finally, the best results are obtained with our method that can exploit cross-task relations more efficiently through joint pairwise task mappings with the proposed regularization.
Interestingly, our method also outperforms the SL baseline that has access to all the task labels, showing the potential information in the cross-task relations.

\begin{table}[h]
	\centering
    \resizebox{0.8\textwidth}{!}
    {
		\begin{tabular}{clccccc}

		    \toprule
		    \# label & Method & Seg. (IoU) $\uparrow$ & Depth (aErr) $\downarrow$ \\
		    \midrule
		    full & Supervised Learning & 73.36 & 0.0165 \\
		    \midrule
		    \multirow{7}{*}{one} & Supervised Learning & 69.50 & 0.0186 \\
		    & Semi-supervised Learning & 71.67 & 0.0178 \\
		    & Perceptual-Map & 72.82 & 0.0169 \\
		    & Direct-Map & 72.33 & 0.0179 \\
		    & Contrastive-Loss & 71.79 & 0.0183 \\
		    & Discriminator-Loss & 68.94 & 0.0208 \\
			\cmidrule{2-4}	
		    & Ours & {\bf 74.90} & {\bf 0.0161} \\
			\bottomrule
		\end{tabular}%
			}
		\vspace{-0.25cm}
		\caption{Multi-task learning results on Cityscapes. `one' indicates each image is randomly annotated with one task label.}
		\label{tab:citys}
\end{table}%
\vspace{-0.2cm}

\paragraph{Results on NYU-v2.}
We then evaluate our method along with the baselines on NYU-v2 in the \emph{random} and \emph{one} label 
settings in \cref{tab:nyuv2}. 
While we observe a similar trend across different methods, overall the performances are lower in this benchmark possibly due to fewer training images than CityScapes.
As expected, the performance in random-label setting is better than the one in one-label setting, as there are more labels available in the former.
While the best results are obtained with SL trained on the full supervision, our method obtains the best performance among the partially supervised methods.
Here SSL improves over SL trained on the partial labels and cross-task consistency is beneficial except for \DM\ in the one label setting, possibly because the dataset is too small to learn accurate mappings between two tasks, while our method is more data-efficient and more successful to exploit the cross-task relations.

\begin{table}[h]
	\centering
	
    \resizebox{0.9\textwidth}{!}
    {
		\begin{tabular}{clcccccccccc}

		    \toprule
		     \# labels & Method & Seg. (IoU) $\uparrow$ & Depth (aErr) $\downarrow$ & Norm. (mErr) $\downarrow$ \\
		    \midrule
		    full & Supervised learning & 36.95 & 0.5510 & 29.51 \\
		    \midrule
		    \multirow{5}{*}{random} & Supervised Learning & 27.05 & 0.6624 & 33.58 \\
		    & Semi-supervised Learning & 29.50 & 0.6224 & 33.31 \\
		    & Perceptual-Map & 32.20 & 0.6037 & 32.07 \\
		    & Direct-Map & 29.17 & 0.6128 & 33.63 \\
		    \cmidrule{2-5}
		    & Ours & {\bf 34.26} & {\bf 0.5787} & {\bf 31.06} \\
		    \midrule
		    \multirow{5}{*}{one} & Supervised Learning & 25.75 & 0.6511 & 33.73 \\
		    & Semi-supervised Learning & 27.52 & 0.6499 & 33.58 \\
		    & Perceptual-Map & 26.94 & 0.6342 & 34.30 \\
		    & Direct-Map & 19.98 & 0.6960 & 37.56 \\
		    \cmidrule{2-5}
		    & Ours & {\bf 30.36} & {\bf 0.6088} & {\bf 32.08} \\
			\bottomrule
		\end{tabular}%
			}
		\vspace{-0.25cm}
		\caption{Multi-task learning results on NYU-v2. `random' indicates each image is annotated with a random number of task labels and `one' means each image is randomly annotated with one task.}
		\label{tab:nyuv2}
\end{table}%
% \vspace{-0.3cm}

\paragraph{Results on PASCAL-Context.}
We evaluate all methods on PASCAL-Context, in both label settings, which contains wider variety of tasks than the previous benchmarks and report the results in \cref{tab:pascal}. 
As the required number of pairwise mappings for \DM\ and \PM\ grows quadratically (20 mappings for 5 tasks), we omit these two due to their high computational cost and compare our method only to SL and SSL baselines.
We see that the SSL baseline improves the performance over SL in random-label setting, however, it performs worse than the SL in one label setting, when there are 60\% less labels.
Again, by exploiting task relations, our method obtains better or comparable results to SSL, while the gains achieved over SL and SSL are more significant in the low label regime (one-label).
Interestingly, SSL and our method obtain comparable results in random-label setting which suggests that relations across tasks are less informative than the ones in CityScape and NYUv2.

\begin{table}[h]
	\centering
	
% 	\resizebox{15cm}{}
    \resizebox{1.0\textwidth}{!}
    {
		\begin{tabular}{clcccccccccc}

		    \toprule
		    \# labels & Method & Seg. (IoU) $\uparrow$ & H. Parts (IoU) $\uparrow$ & Norm. (mErr) $\downarrow$ & Sal. (IoU) $\uparrow$ & Edge (odsF) $\uparrow$ \\
		    \midrule
		    full & Supervised Learning & 63.9 & 58.9 & 15.1 & 65.4 & 69.4 \\
		    \midrule
		    \multirow{3}{*}{random} 
		    & Supervised Learning & 58.4 & 55.3 & 16.0 & 63.9 & {\bf 67.8} \\
		    & Semi-supervised Learning & {\bf 59.0} & {\bf 55.8} & {\bf 15.9} & {\bf 64.0} & 66.9 \\
		    & Ours & {\bf 59.0} & 55.6 & {\bf 15.9} & {\bf 64.0} & {\bf 67.8} \\
		    \midrule
		    \multirow{3}{*}{one}
		    & Supervised Learning & 48.0 & 55.6 & 17.2 & 61.5 & 64.6 \\
		    & Semi-supervised Learning & 45.0 & 54.0 & {\bf 16.9} & {\bf 61.7} & 62.4 \\
		    & Ours & {\bf 49.5} & {\bf 55.8} & 17.0 & {\bf 61.7} & {\bf 65.1} \\
			\bottomrule
		\end{tabular}%
			}
		\vspace{-0.25cm}
		\caption{Multi-task learning results on PASCAL. `random' indicates each image is annotated with a random number of task labels and `one' means each image is randomly annotated with one task.}
		\label{tab:pascal}
\end{table}%
\vspace{-0.2cm}

\subsection{Further results}

\paragraph{Learning from partial and imbalanced task labels.}
So far, we considered the partially annotated setting where the number of labels for each task is similar. We further evaluate all methods in an imbalanced partially supervised setting in Cityscapes, where we assume the ratio of labels for each task are imbalanced, \eg we randomly sample 90\% of images to be labeled for semantic segmentation and only 10\% images having labels for depth and we denote this setting by the label ratio between segmentation and depth (Seg.:Depth = 9:1). 
The opposite case (Seg.:Depth = 1:9) is also considered.

\begin{table}[h]
	\centering
	
% 	\resizebox{15cm}{}
    \resizebox{0.8\textwidth}{!}
    {
		\begin{tabular}{clccccc}

		    \toprule
		    \#labels & Method & Seg. (IoU) $\uparrow$ & Depth (aErr) $\downarrow$ \\
		    \midrule
		    full & Supervised Learning & 73.36 & 0.0165 \\
		    \midrule
		    \multirow{5}{*}{1:9} & Supervised Learning & 63.37 & 0.0161 \\
		    & Semi-supervised Learning & 64.40 & 0.0179 \\
		    & Perceptual-Map & 68.84 & 0.0141 \\
		    & Direct-Map & 67.04 & 0.0153 \\
		    \cmidrule{2-4}
		    & Ours & {\bf 71.89} & {\bf 0.0131} \\
		    \midrule
		    \multirow{5}{*}{9:1} & Supervised learning & 72.77 & 0.0250 \\
		    & Semi-supervised Learning & 72.97 & 0.0395 \\
		    & Perceptual-Map & 73.36 & 0.0237 \\
		    & Direct-Map & 73.13 & 0.0288 \\
		    \cmidrule{2-4}
		    & Ours & {\bf 74.23} & {\bf 0.0235} \\
			\bottomrule
		\end{tabular}%
			}
		\vspace{-0.25cm}
		\caption{Multi-task learning results on Cityscapes. `\#label' indicates the number ratio of labels for segmentation and depth, \eg `1:9' means we have 10\% of images annotated with segmentation labels and 90\% of images have depth groundtruth.}
		\label{tab:citysimbalance}
\end{table}%

We report the results in \cref{tab:citysimbalance}.
The performance of supervised learning (SL) on the task with partial labels drops significantly. Though SSL improves the performance on segmentation, its performance on depth drops in both cases. 
In contrast to SL and SSL, our method and Perceptual-Map obtain better results on all tasks in both settings by learning cross-task consistency while our method obtains the best results by joint space mapping. 
This demonstrates that our model can successfully learn cross-task relations from unbalanced labels thanks to its task agnostic mapping function which can share parameters across multiple task pairs.

\paragraph{Cross-task consistency learning with full supervision.}
Our method can also be applied to fully-supervised learning setting where all task labels are available for each sample by mapping one task's prediction and another task's ground-truth to the joint space and measuring cross-task consistency in the joint space. We applied our method to NYU-v2 and compare it with the single task learning (STL) networks, vanilla MTL baseline, recent multi-task learning methods, \ie MTAN~\cite{liu2019end}, X-task~\cite{zamir2020robust}, and several methods focusing on loss weighting strategies, \ie Uncertainty~\cite{kendall2018multi}, GradNorm~\cite{chen2018gradnorm}, MGDA~\cite{sener2018multi} and DWA~\cite{liu2019end} in \cref{tab:nyuv2full}. 

\begin{table}[h!]
	\centering
	
% 	\resizebox{15cm}{}
    \resizebox{0.8\textwidth}{!}
    {
		\begin{tabular}{lccccccccccc}

		    \toprule
		    Method & Seg. (IoU) $\uparrow$ & Depth (aErr) $\downarrow$ & Norm. (mErr) $\downarrow$ \\
		    \midrule
		    STL & 37.45 & 0.6079 & 25.94 \\
		    \midrule
		    MTL & 36.95 & 0.5510 & 29.51 \\
		    MTAN~\cite{liu2019end} & 39.39 & 0.5696 & 28.89 \\
		    X-task~\cite{zamir2020robust} & 38.91 &  0.5342 & 29.94 \\
		    Uncertainty~\cite{kendall2018multi} & 36.46 & 0.5376 & 27.58 \\
		    GradNorm~\cite{chen2018gradnorm} & 37.19 & 0.5775 & 28.51 \\
		    MGDA~\cite{sener2018multi} & 38.65 & 0.5572 & 28.89 \\
		    DWA~\cite{liu2019end} & 36.46 & 0.5429 & 29.45 \\
		    \midrule
		    Ours & 41.00 & 0.5148 & 28.58 \\
		    Ours + Uncertainty & {\bf 41.09} & {\bf 0.5090} & {\bf 26.78} \\
			\bottomrule
		\end{tabular}%
			}
		\vspace{-0.25cm}
		\caption{Multi-task fully-supervised learning results on NYU-v2. `STL' indicates standard single-task learning and `MTL' means the standard multi-task learning network.}
		\label{tab:nyuv2full}
\end{table}%

MTL, MTAN, X-task and Ours are trained with uniform loss weights. 
We see that our method (Ours) performs better than the other methods with uniform loss weights, \eg MTAN and X-task, where X-task regularizes cross-task consistency by learning perceptual loss with pre-trained cross-task mapping functions. 
This shows that cross-task consistency is informative even in the fully supervised case and our method is more effective for learning cross-task consistency. 
Compared to recent loss weighting strategies, our method (Ours) obtains better performance on segmentation and depth estimation than other methods while slightly worse on normal estimation compared with GradNorm and Uncertainty. 
This is because the loss weighting strategies enable a more balanced optimization of multi-task learning than uniformly loss weighting. 
Thus when we incorporate the loss weighing strategy of Uncertainty~\cite{kendall2018multi} to our method, \ie (Ours + Uncertainty), our method obtains further improvement and outperforms both GradNorm and Uncertainty.

% Ablation study
\subsection{Ablation study}
Here, we conduct an ablation study to evaluate the effect of task-pair conditional mapping function and the regularization in \cref{eq:mtlours}. 
To this end, we report results of our method without task-pair condition network ($a_{\theta}$), denoted as `Ours (w/o cond)' where we use a single mapping  ($\bar{m}_{\vartheta}$) for all task pairs, and also our method without the regularization in \cref{eq:mtlours}, denoted as `Ours (w/o reg)' in \cref{tab:nyuv2abla}. 
First our full model outperforms both Ours (w/o cond) and Ours (w/o reg) which shows that both the components are beneficial.
Ours (w/o cond) which employs the same mapping for all the task pairs still achieves better performance than the SL baseline.
Surprisingly, even after removing the regularization, despite the performance drop, the pairwise mappings can still be regulated with a lower learning rate to avoid learning trivial mappings and it still outperforms the SL baseline.

\begin{table}[h!]
	\centering
	
% 	\resizebox{15cm}{}
    \resizebox{1.0\textwidth}{!}
    {
		\begin{tabular}{clcccccccccc}

		    \toprule
		    \# labels & Method & Seg. (IoU) $\uparrow$ & Depth (aErr) $\downarrow$ & Norm. (mErr) $\downarrow$ \\
		    \midrule
		    \multirow{4}{*}{random} 
		    & Supervised Learning & 27.05 & 0.6624 & 33.58 \\
		    & Ours (w/o cond) & 34.13 & 0.5968 & 31.65 \\
		    & Ours (w/o reg) & 33.87 & 0.5887 & 31.24 \\
		    & Ours & {\bf 34.26} & {\bf 0.5787} & {\bf 31.06} \\
		    \midrule
		    \multirow{4}{*}{one} 
		    & Supervised Learning & 25.75 & 0.6511 & 33.73 \\
		    & Ours (w/o cond) & 29.19 & 0.6181 & 32.62 \\
		    & Ours (w/o reg) & 28.36 & 0.6407 & 32.92 \\
		    & Ours & {\bf 30.36} & {\bf 0.6088} & {\bf 32.08} \\ 
			\bottomrule
		\end{tabular}%
			}
		\vspace{-0.25cm}
		\caption{Ablation study on NYU-v2. `cond' indicates whether using conditional mapping function. `reg' indicates whether we use regularization in \cref{eq:mtlours}.}
		\label{tab:nyuv2abla}
\end{table}%
\vspace{-0.2cm}

\subsection{Qualitative results}
\begin{figure}[h!]
\begin{center}
\includegraphics[width=0.8\linewidth]{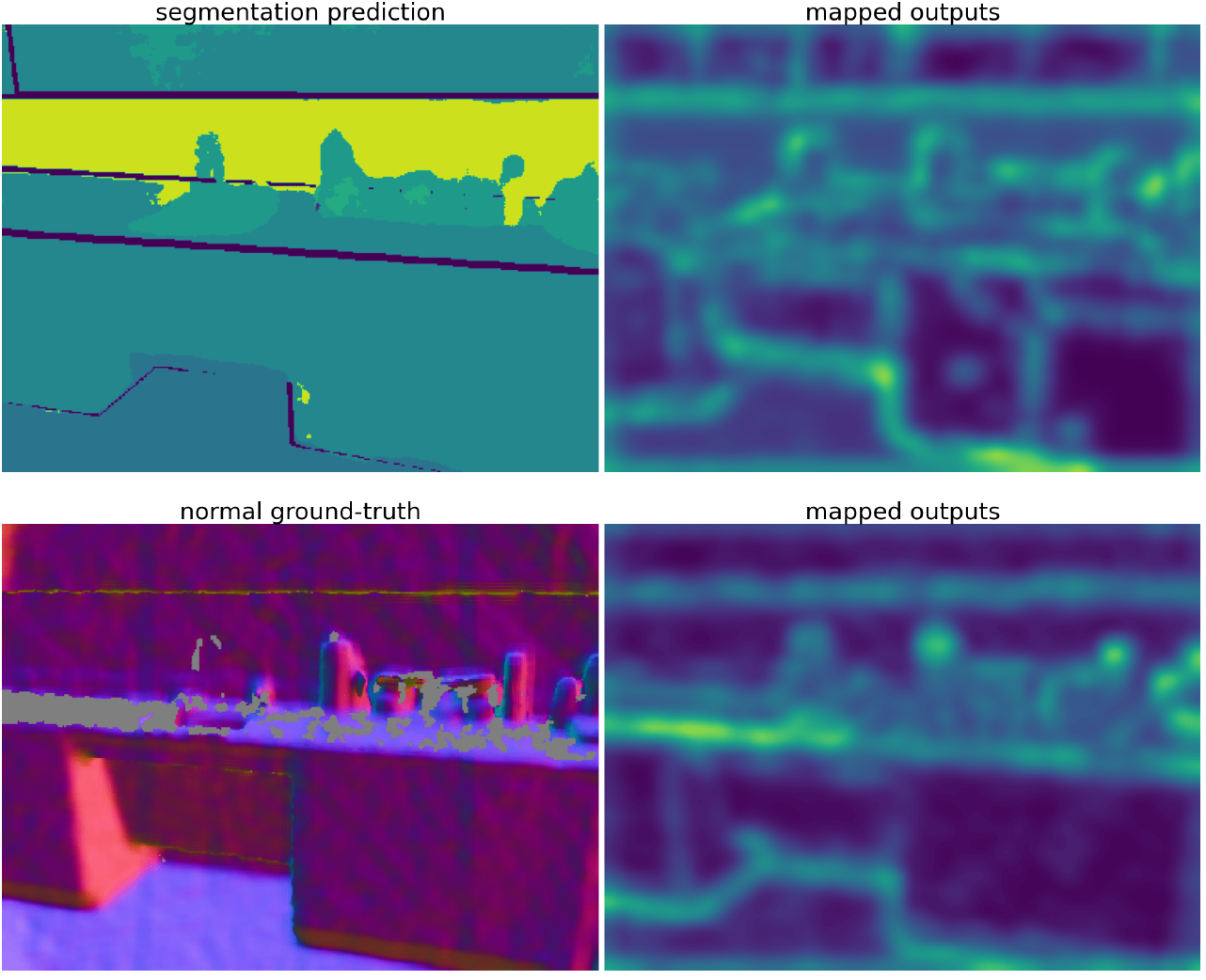}
\end{center}
\vspace{-0.65cm}
\caption{Intermediate feature map of the mapping function of the task-pair (segmentation to surface normal) of one example in NYU-v2. The first column shows the prediction or ground-truth and the second column present the corresponding mapped feature map (output of the mapping function's last second layer ).}
\label{fig:mapped}
\end{figure}

\vspace{-0.1cm}
Here, we present some qualitative results and refer to the supplementary for more results.
\paragraph{Mapped outputs.}
Here, we visualize the intermediate feature maps of $m^{s\rightarrow st}$ and $m^{t\rightarrow st}$  for one example in NYU-v2 in \cref{fig:mapped} where $s$ and $t$ correspond to segmentation and surface normal estimation respectively.
We observe that the functions map both task labels to a joint pairwise space where the common information is around object boundaries, which in turn enables the model to produce more accurate predictions for both tasks.

% \vspace{-0.2cm}

\begin{figure}[h!]
\begin{center}
\includegraphics[width=1.0\linewidth]{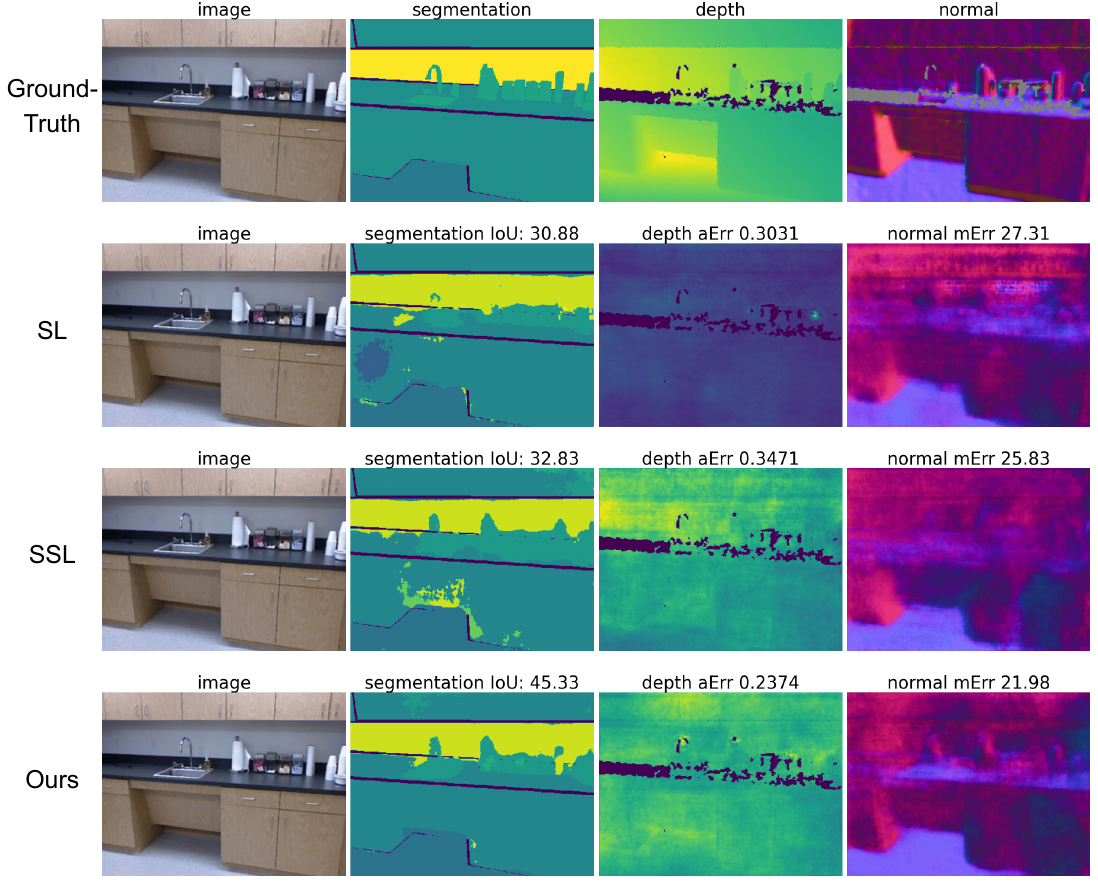}
\end{center}
\vspace{-0.65cm}
\caption{\textbf{Qualitative results on NYU-v2.} The fist column shows the RGB image, the second column plots the ground-truth or predictions with the IoU ($\uparrow$) score of all methods for semantic segmentation, the third column presents the ground-truth or predictions with the absolute error ($\downarrow$), and we show the prediction of surface normal with mean error ($\downarrow$) in the last column.}
\label{fig:predictions}
\end{figure}

\paragraph{Predictions.}
Finally we show qualitative comparisons between our method, SL and SSL baselines on NYU-v2 in \cref{fig:predictions}. 
We can see that our method produces more accurate predictions by leveraging cross-task consistency. We also provide additional experiments in supplementary.

\section{Conclusion and Limitations}\label{sec:con}
% !TEX root = main.tex
\vspace{-0.1cm}

In this paper, we show that cross-task relations are crucial to learn multi-task dense prediction problems from partially annotated data in several benchmarks.
We present a model agnostic method that learns relations between task pairs in joint latent spaces through mapping functions conditioned on the task pair in a computationally efficient way and also avoids learning trivial mappings with a regularization strategy.
Finally, our method has limitations too. Despite the efficient learning of cross-task relations through a conditioned network, modeling cross-task relations for all task pairs may not be required. Thus it would be desirable to automatically identify which tasks are closely related and only learn such cross-task relations.

\paragraph{Acknowledgments.} HB is supported by the EPSRC programme grant Visual AI EP/T028572/1.

{\small
\bibliographystyle{ieee_fullname}
\bibliography{ref}

\begin{thebibliography}{10}\itemsep=-1pt

\bibitem{badrinarayanan2017segnet}
Vijay Badrinarayanan, Alex Kendall, and Roberto Cipolla.
\newblock Segnet: A deep convolutional encoder-decoder architecture for image
  segmentation.
\newblock {\em PAMI}, 39(12):2481--2495, 2017.

\bibitem{berthelot2019mixmatch}
David Berthelot, Nicholas Carlini, Ian Goodfellow, Nicolas Papernot, Avital
  Oliver, and Colin Raffel.
\newblock Mixmatch: A holistic approach to semi-supervised learning.
\newblock {\em NeurIPS}, 2019.

\bibitem{bilen2016integrated}
Hakan Bilen and Andrea Vedaldi.
\newblock Integrated perception with recurrent multi-task neural networks.
\newblock In {\em Advances in Neural Information Processing Systems}, pages
  235--243, 2016.

\bibitem{bragman2019stochastic}
Felix~JS Bragman, Ryutaro Tanno, Sebastien Ourselin, Daniel~C Alexander, and
  Jorge Cardoso.
\newblock Stochastic filter groups for multi-task cnns: Learning specialist and
  generalist convolution kernels.
\newblock In {\em ICCV}, pages 1385--1394, 2019.

\bibitem{bruggemann2020automated}
David Bruggemann, Menelaos Kanakis, Stamatios Georgoulis, and Luc Van~Gool.
\newblock Automated search for resource-efficient branched multi-task networks.
\newblock {\em arXiv preprint arXiv:2008.10292}, 2020.

\bibitem{bruggemann2021exploring}
David Bruggemann, Menelaos Kanakis, Anton Obukhov, Stamatios Georgoulis, and
  Luc Van~Gool.
\newblock Exploring relational context for multi-task dense prediction.
\newblock In {\em ICCV}, 2021.

\bibitem{caruana1997multitask}
Rich Caruana.
\newblock Multitask learning.
\newblock {\em Machine learning}, 28(1):41--75, 1997.

\bibitem{casser2019depth}
Vincent Casser, Soeren Pirk, Reza Mahjourian, and Anelia Angelova.
\newblock Depth prediction without the sensors: Leveraging structure for
  unsupervised learning from monocular videos.
\newblock In {\em AAAI}, volume~33, pages 8001--8008, 2019.

\bibitem{chen2017deeplab}
Liang-Chieh Chen, George Papandreou, Iasonas Kokkinos, Kevin Murphy, and Alan~L
  Yuille.
\newblock Deeplab: Semantic image segmentation with deep convolutional nets,
  atrous convolution, and fully connected crfs.
\newblock {\em PAMI}, 40(4):834--848, 2017.

\bibitem{chen2018encoder}
Liang-Chieh Chen, Yukun Zhu, George Papandreou, Florian Schroff, and Hartwig
  Adam.
\newblock Encoder-decoder with atrous separable convolution for semantic image
  segmentation.
\newblock In {\em ECCV}, pages 801--818, 2018.

\bibitem{chen2014detect}
Xianjie Chen, Roozbeh Mottaghi, Xiaobai Liu, Sanja Fidler, Raquel Urtasun, and
  Alan Yuille.
\newblock Detect what you can: Detecting and representing objects using
  holistic models and body parts.
\newblock In {\em CVPR}, pages 1971--1978, 2014.

\bibitem{chen2018gradnorm}
Zhao Chen, Vijay Badrinarayanan, Chen-Yu Lee, and Andrew Rabinovich.
\newblock Gradnorm: Gradient normalization for adaptive loss balancing in deep
  multitask networks.
\newblock In {\em ICML}, pages 794--803. PMLR, 2018.

\bibitem{chen2020just}
Zhao Chen, Jiquan Ngiam, Yanping Huang, Thang Luong, Henrik Kretzschmar, Yuning
  Chai, and Dragomir Anguelov.
\newblock Just pick a sign: Optimizing deep multitask models with gradient sign
  dropout.
\newblock {\em NeurIPS}, 2020.

\bibitem{chen2020multi}
Zhihao Chen, Lei Zhu, Liang Wan, Song Wang, Wei Feng, and Pheng-Ann Heng.
\newblock A multi-task mean teacher for semi-supervised shadow detection.
\newblock In {\em CVPR}, pages 5611--5620, 2020.

\bibitem{cordts2016cityscapes}
Marius Cordts, Mohamed Omran, Sebastian Ramos, Timo Rehfeld, Markus Enzweiler,
  Rodrigo Benenson, Uwe Franke, Stefan Roth, and Bernt Schiele.
\newblock The cityscapes dataset for semantic urban scene understanding.
\newblock In {\em Computer Vision and Pattern Recognition}, pages 3213--3223,
  2016.

\bibitem{eigen2015predicting}
David Eigen and Rob Fergus.
\newblock Predicting depth, surface normals and semantic labels with a common
  multi-scale convolutional architecture.
\newblock In {\em IEEE International Conference on Computer Vision}, pages
  2650--2658, 2015.

\bibitem{eigen2014depth}
David Eigen, Christian Puhrsch, and Rob Fergus.
\newblock Depth map prediction from a single image using a multi-scale deep
  network.
\newblock {\em arXiv preprint arXiv:1406.2283}, 2014.

\bibitem{everingham2010pascal}
Mark Everingham, Luc Van~Gool, Christopher~KI Williams, John Winn, and Andrew
  Zisserman.
\newblock The pascal visual object classes (voc) challenge.
\newblock {\em IJCV}, 88(2):303--338, 2010.

\bibitem{gao2019nddr}
Yuan Gao, Jiayi Ma, Mingbo Zhao, Wei Liu, and Alan~L Yuille.
\newblock Nddr-cnn: Layerwise feature fusing in multi-task cnns by neural
  discriminative dimensionality reduction.
\newblock In {\em CVPR}, pages 3205--3214, 2019.

\bibitem{geiger2012we}
Andreas Geiger, Philip Lenz, and Raquel Urtasun.
\newblock Are we ready for autonomous driving? the kitti vision benchmark
  suite.
\newblock In {\em CVPR}, pages 3354--3361. IEEE, 2012.

\bibitem{gong2019comparison}
Ting Gong, Tyler Lee, Cory Stephenson, Venkata Renduchintala, Suchismita Padhy,
  Anthony Ndirango, Gokce Keskin, and Oguz~H Elibol.
\newblock A comparison of loss weighting strategies for multi task learning in
  deep neural networks.
\newblock {\em IEEE Access}, 7:141627--141632, 2019.

\bibitem{guizilini2020robust}
Vitor Guizilini, Jie Li, Rares Ambrus, Sudeep Pillai, and Adrien Gaidon.
\newblock Robust semi-supervised monocular depth estimation with reprojected
  distances.
\newblock In {\em Conference on robot learning}, pages 503--512. PMLR, 2020.

\bibitem{guo2018dynamic}
Michelle Guo, Albert Haque, De-An Huang, Serena Yeung, and Li Fei-Fei.
\newblock Dynamic task prioritization for multitask learning.
\newblock In {\em ECCV}, pages 270--287, 2018.

\bibitem{guo2020learning}
Pengsheng Guo, Chen-Yu Lee, and Daniel Ulbricht.
\newblock Learning to branch for multi-task learning.
\newblock In {\em ICML}, pages 3854--3863. PMLR, 2020.

\bibitem{he2017mask}
Kaiming He, Georgia Gkioxari, Piotr Doll{\'a}r, and Ross Girshick.
\newblock Mask r-cnn.
\newblock In {\em ICCV}, pages 2961--2969, 2017.

\bibitem{he2016deep}
Kaiming He, Xiangyu Zhang, Shaoqing Ren, and Jian Sun.
\newblock Deep residual learning for image recognition.
\newblock In {\em CVPR}, pages 770--778, 2016.

\bibitem{hoiem2008closing}
Derek Hoiem, Alexei~A Efros, and Martial Hebert.
\newblock Closing the loop in scene interpretation.
\newblock In {\em CVPR}, pages 1--8. IEEE, 2008.

\bibitem{hoyer2021improving}
Lukas Hoyer, Dengxin Dai, Qin Wang, Yuhua Chen, and Luc Van~Gool.
\newblock Improving semi-supervised and domain-adaptive semantic segmentation
  with self-supervised depth estimation.
\newblock {\em arXiv preprint arXiv:2108.12545}, 2021.

\bibitem{imran2020partly}
Abdullah-Al-Zubaer Imran, Chao Huang, Hui Tang, Wei Fan, Yuan Xiao, Dingjun
  Hao, Zhen Qian, and Demetri Terzopoulos.
\newblock Partly supervised multitask learning.
\newblock {\em arXiv preprint arXiv:2005.02523}, 2020.

\bibitem{kendall2018multi}
Alex Kendall, Yarin Gal, and Roberto Cipolla.
\newblock Multi-task learning using uncertainty to weigh losses for scene
  geometry and semantics.
\newblock In {\em CVPR}, pages 7482--7491, 2018.

\bibitem{kuznietsov2017semi}
Yevhen Kuznietsov, Jorg Stuckler, and Bastian Leibe.
\newblock Semi-supervised deep learning for monocular depth map prediction.
\newblock In {\em CVPR}, pages 6647--6655, 2017.

\bibitem{latif2019multi}
Siddique Latif, Rajib Rana, Sara Khalifa, Raja Jurdak, Julien Epps, and
  Bj{\"o}rn~W Schuller.
\newblock Multi-task semi-supervised adversarial autoencoding for speech
  emotion recognition.
\newblock {\em arXiv preprint arXiv:1907.06078}, 2019.

\bibitem{li2020knowledge}
Wei-Hong Li and Hakan Bilen.
\newblock Knowledge distillation for multi-task learning.
\newblock In {\em ECCV Workshop on Imbalance Problems in Computer Vision},
  pages 163--176. Springer, 2020.

\bibitem{lin2019pareto}
Xi Lin, Hui-Ling Zhen, Zhenhua Li, Qing-Fu Zhang, and Sam Kwong.
\newblock Pareto multi-task learning.
\newblock {\em NeurIPS}, 32:12060--12070, 2019.

\bibitem{liu2010single}
Beyang Liu, Stephen Gould, and Daphne Koller.
\newblock Single image depth estimation from predicted semantic labels.
\newblock In {\em CVPR}, pages 1253--1260. IEEE, 2010.

\bibitem{liu2008semi}
Qiuhua Liu, Xuejun Liao, and Lawrence Carin.
\newblock Semi-supervised multitask learning.
\newblock In {\em Advances in Neural Information Processing Systems}, pages
  937--944, 2008.

\bibitem{liu2019end}
Shikun Liu, Edward Johns, and Andrew~J Davison.
\newblock End-to-end multi-task learning with attention.
\newblock In {\em CVPR}, pages 1871--1880, 2019.

\bibitem{long2015fully}
Jonathan Long, Evan Shelhamer, and Trevor Darrell.
\newblock Fully convolutional networks for semantic segmentation.
\newblock In {\em CVPR}, pages 3431--3440, 2015.

\bibitem{lu2017fully}
Yongxi Lu, Abhishek Kumar, Shuangfei Zhai, Yu Cheng, Tara Javidi, and Rogerio
  Feris.
\newblock Fully-adaptive feature sharing in multi-task networks with
  applications in person attribute classification.
\newblock In {\em CVPR}, pages 5334--5343, 2017.

\bibitem{lu2021taskology}
Yao Lu, Soren Pirk, Jan Dlabal, Anthony Brohan, Ankita Pasad, Zhao Chen,
  Vincent Casser, Anelia Angelova, and Ariel Gordon.
\newblock Taskology: Utilizing task relations at scale.
\newblock In {\em CVPR}, pages 8700--8709, 2021.

\bibitem{martin2004learning}
David~R Martin, Charless~C Fowlkes, and Jitendra Malik.
\newblock Learning to detect natural image boundaries using local brightness,
  color, and texture cues.
\newblock {\em PAMI}, 26(5):530--549, 2004.

\bibitem{mendel2020semi}
Robert Mendel, Luis~Antonio De~Souza, David Rauber, Jo{\~a}o~Paulo Papa, and
  Christoph Palm.
\newblock Semi-supervised segmentation based on error-correcting supervision.
\newblock In {\em European Conference on Computer Vision}, pages 141--157.
  Springer, 2020.

\bibitem{minaee2021image}
Shervin Minaee, Yuri~Y Boykov, Fatih Porikli, Antonio~J Plaza, Nasser
  Kehtarnavaz, and Demetri Terzopoulos.
\newblock Image segmentation using deep learning: A survey.
\newblock {\em PAMI}, 2021.

\bibitem{misra2016cross}
Ishan Misra, Abhinav Shrivastava, Abhinav Gupta, and Martial Hebert.
\newblock Cross-stitch networks for multi-task learning.
\newblock In {\em CVPR}, pages 3994--4003, 2016.

\bibitem{olsson2021classmix}
Viktor Olsson, Wilhelm Tranheden, Juliano Pinto, and Lennart Svensson.
\newblock Classmix: Segmentation-based data augmentation for semi-supervised
  learning.
\newblock In {\em WACV}, pages 1369--1378, 2021.

\bibitem{perez2018film}
Ethan Perez, Florian Strub, Harm De~Vries, Vincent Dumoulin, and Aaron
  Courville.
\newblock Film: Visual reasoning with a general conditioning layer.
\newblock In {\em Proceedings of the AAAI Conference on Artificial
  Intelligence}, 2018.

\bibitem{poggi2020uncertainty}
Matteo Poggi, Filippo Aleotti, Fabio Tosi, and Stefano Mattoccia.
\newblock On the uncertainty of self-supervised monocular depth estimation.
\newblock In {\em CVPR}, pages 3227--3237, 2020.

\bibitem{ruder2017overview}
Sebastian Ruder.
\newblock An overview of multi-task learning in deep neural networks.
\newblock {\em arXiv preprint arXiv:1706.05098}, 2017.

\bibitem{ruder2019latent}
Sebastian Ruder, Joachim Bingel, Isabelle Augenstein, and Anders S{\o}gaard.
\newblock Latent multi-task architecture learning.
\newblock In {\em AAAI}, volume~33, pages 4822--4829, 2019.

\bibitem{saha2021learning}
Suman Saha, Anton Obukhov, Danda~Pani Paudel, Menelaos Kanakis, Yuhua Chen,
  Stamatios Georgoulis, and Luc Van~Gool.
\newblock Learning to relate depth and semantics for unsupervised domain
  adaptation.
\newblock In {\em CVPR}, pages 8197--8207, 2021.

\bibitem{sener2018multi}
Ozan Sener and Vladlen Koltun.
\newblock Multi-task learning as multi-objective optimization.
\newblock {\em NeurIPS}, 2018.

\bibitem{silberman2012indoor}
Nathan Silberman, Derek Hoiem, Pushmeet Kohli, and Rob Fergus.
\newblock Indoor segmentation and support inference from rgbd images.
\newblock In {\em European conference on computer vision}, pages 746--760.
  Springer, 2012.

\bibitem{sohn2020fixmatch}
Kihyuk Sohn, David Berthelot, Chun-Liang Li, Zizhao Zhang, Nicholas Carlini,
  Ekin~D Cubuk, Alex Kurakin, Han Zhang, and Colin Raffel.
\newblock Fixmatch: Simplifying semi-supervised learning with consistency and
  confidence.
\newblock {\em NeurIPS}, 2020.

\bibitem{su2021pixel}
Zhuo Su, Wenzhe Liu, Zitong Yu, Dewen Hu, Qing Liao, Qi Tian, Matti
  Pietikainen, and Li Liu.
\newblock Pixel difference networks for efficient edge detection.
\newblock In {\em Proceedings of the IEEE/CVF International Conference on
  Computer Vision}, pages 5117--5127, 2021.

\bibitem{sun2021see}
Boyang Sun, Jiaxu Xing, Hermann Blum, Roland Siegwart, and Cesar Cadena.
\newblock See yourself in others: Attending multiple tasks for own failure
  detection.
\newblock {\em arXiv preprint arXiv:2110.02549}, 2021.

\bibitem{terzopoulos2019semi}
Demetri Terzopoulos et~al.
\newblock Semi-supervised multi-task learning with chest x-ray images.
\newblock In {\em International Workshop on Machine Learning in Medical
  Imaging}, pages 151--159. Springer, 2019.

\bibitem{tian2020conditional}
Zhi Tian, Chunhua Shen, and Hao Chen.
\newblock Conditional convolutions for instance segmentation.
\newblock In {\em ECCV}, pages 282--298. Springer, 2020.

\bibitem{vandenhende2019branched}
Simon Vandenhende, Stamatios Georgoulis, Bert De~Brabandere, and Luc Van~Gool.
\newblock Branched multi-task networks: deciding what layers to share.
\newblock In {\em BMVC}, 2020.

\bibitem{vandenhende2021multi}
Simon Vandenhende, Stamatios Georgoulis, Wouter Van~Gansbeke, Marc Proesmans,
  Dengxin Dai, and Luc Van~Gool.
\newblock Multi-task learning for dense prediction tasks: A survey.
\newblock {\em PAMI}, 2021.

\bibitem{vandenhende2020mti}
Simon Vandenhende, Stamatios Georgoulis, and Luc Van~Gool.
\newblock Mti-net: Multi-scale task interaction networks for multi-task
  learning.
\newblock In {\em ECCV}, pages 527--543. Springer, 2020.

\bibitem{voges2018timestamp}
Raphael Voges and Bernardo Wagner.
\newblock Timestamp offset calibration for an imu-camera system under interval
  uncertainty.
\newblock In {\em 2018 IEEE/RSJ International Conference on Intelligent Robots
  and Systems (IROS)}, pages 377--384. IEEE, 2018.

\bibitem{wang2009semi}
Fei Wang, Xin Wang, and Tao Li.
\newblock Semi-supervised multi-task learning with task regularizations.
\newblock In {\em ICDM}, pages 562--568. IEEE, 2009.

\bibitem{wang2021domain}
Qin Wang, Dengxin Dai, Lukas Hoyer, Luc Van~Gool, and Olga Fink.
\newblock Domain adaptive semantic segmentation with self-supervised depth
  estimation.
\newblock In {\em ICCV}, pages 8515--8525, 2021.

\bibitem{wang2021end}
Yuqing Wang, Zhaoliang Xu, Xinlong Wang, Chunhua Shen, Baoshan Cheng, Hao Shen,
  and Huaxia Xia.
\newblock End-to-end video instance segmentation with transformers.
\newblock In {\em CVPR}, pages 8741--8750, 2021.

\bibitem{xia2017joint}
Fangting Xia, Peng Wang, Xianjie Chen, and Alan~L Yuille.
\newblock Joint multi-person pose estimation and semantic part segmentation.
\newblock In {\em CVPR}, pages 6769--6778, 2017.

\bibitem{xu2018pad}
Dan Xu, Wanli Ouyang, Xiaogang Wang, and Nicu Sebe.
\newblock Pad-net: Multi-tasks guided prediction-and-distillation network for
  simultaneous depth estimation and scene parsing.
\newblock In {\em CVPR}, pages 675--684, 2018.

\bibitem{yu2020gradient}
Tianhe Yu, Saurabh Kumar, Abhishek Gupta, Sergey Levine, Karol Hausman, and
  Chelsea Finn.
\newblock Gradient surgery for multi-task learning.
\newblock {\em NeurIPS}, 2020.

\bibitem{yu2017casenet}
Zhiding Yu, Chen Feng, Ming-Yu Liu, and Srikumar Ramalingam.
\newblock Casenet: Deep category-aware semantic edge detection.
\newblock In {\em CVPR}, pages 5964--5973, 2017.

\bibitem{zamir2020robust}
Amir~R Zamir, Alexander Sax, Nikhil Cheerla, Rohan Suri, Zhangjie Cao, Jitendra
  Malik, and Leonidas~J Guibas.
\newblock Robust learning through cross-task consistency.
\newblock In {\em CVPR}, pages 11197--11206, 2020.

\bibitem{zamir2018taskonomy}
Amir~R Zamir, Alexander Sax, William Shen, Leonidas~J Guibas, Jitendra Malik,
  and Silvio Savarese.
\newblock Taskonomy: Disentangling task transfer learning.
\newblock In {\em CVPR}, pages 3712--3722, 2018.

\bibitem{zamir2016generic}
Amir~R Zamir, Tilman Wekel, Pulkit Agrawal, Colin Wei, Jitendra Malik, and
  Silvio Savarese.
\newblock Generic 3d representation via pose estimation and matching.
\newblock In {\em ECCV}, pages 535--553. Springer, 2016.

\bibitem{zhang2020uc}
Jing Zhang, Deng-Ping Fan, Yuchao Dai, Saeed Anwar, Fatemeh~Sadat Saleh, Tong
  Zhang, and Nick Barnes.
\newblock Uc-net: Uncertainty inspired rgb-d saliency detection via conditional
  variational autoencoders.
\newblock In {\em CVPR}, pages 8582--8591, 2020.

\bibitem{zhang2020select}
Miao Zhang, Weisong Ren, Yongri Piao, Zhengkun Rong, and Huchuan Lu.
\newblock Select, supplement and focus for rgb-d saliency detection.
\newblock In {\em CVPR}, pages 3472--3481, 2020.

\bibitem{zhang2017survey}
Yu Zhang and Qiang Yang.
\newblock A survey on multi-task learning.
\newblock {\em arXiv preprint arXiv:1707.08114}, 2017.

\bibitem{zhang2009semi}
Yu Zhang and Dit-Yan Yeung.
\newblock Semi-supervised multi-task regression.
\newblock In {\em ECML PKDD}, pages 617--631. Springer, 2009.

\bibitem{zhang2018joint}
Zhenyu Zhang, Zhen Cui, Chunyan Xu, Zequn Jie, Xiang Li, and Jian Yang.
\newblock Joint task-recursive learning for semantic segmentation and depth
  estimation.
\newblock In {\em ECCV}, pages 235--251, 2018.

\bibitem{zhang2019pattern}
Zhenyu Zhang, Zhen Cui, Chunyan Xu, Yan Yan, Nicu Sebe, and Jian Yang.
\newblock Pattern-affinitive propagation across depth, surface normal and
  semantic segmentation.
\newblock In {\em CVPR}, pages 4106--4115, 2019.

\bibitem{zhou2020pattern}
Ling Zhou, Zhen Cui, Chunyan Xu, Zhenyu Zhang, Chaoqun Wang, Tong Zhang, and
  Jian Yang.
\newblock Pattern-structure diffusion for multi-task learning.
\newblock In {\em CVPR}, pages 4514--4523, 2020.

\bibitem{zhou2017unsupervised}
Tinghui Zhou, Matthew Brown, Noah Snavely, and David~G Lowe.
\newblock Unsupervised learning of depth and ego-motion from video.
\newblock In {\em CVPR}, pages 1851--1858, 2017.

\end{thebibliography}
}

\clearpage
\appendix
% !TEX root = suppmain.tex

\section{Implementation Details}
\cref{supptab:datasets} and \cref{supptab:psl} provide an overview of the experimental settings, in particular report the number of train and test samples for each benchmark and number of labels used in different partially annotated settings respectively.
Next we explain the implementation details for each dataset.

\paragraph{Cityscapes.} The Cityscapes dataset~\cite{cordts2016cityscapes} contains 3475 labelled images. 
As in~\cite{liu2019end}, we use 2975 images for training and 500 images for testing. 
In multi-task partially supervised learning setting, we consider the one-label setting in Cityscapes, as there are only two tasks in total, \ie we randomly select and keep label only for 1 task for each training image, resulting in 1487 training images annotated for segmentation and 1488 training images labelled for depth estimation, as shown in \cref{supptab:psl}.

We follow the training and evaluation protocol in~\cite{liu2019end} and we use SegNet~\cite{badrinarayanan2017segnet} as the MTL backbone for all methods, use cross-entropy loss for semantic segmentation, l1-norm loss for depth estimation. 
We use the exactly same hyper-parameters including learning rate, optimizer as in~\cite{liu2019end}. 
More specifically, we use Adam optimizer with a learning rate of 0.0001 and train all models for 200 epochs with a batch size of 8 and halve the learning rate at the 100-th epoch. 
We also employ the same evaluation metrics, mean intersection over union (mIoU) and absolute error (aErr) to evaluate the semantic segmentation and depth estimation task, respectively as in~\cite{liu2019end}. 

For our model, we use the encoder architecture of SegNet for instantiating the joint pairwise task mapping ($\bar{m}_{\vartheta}$) and include one convolutional layer as task specific input layer in $\bar{m}_{\vartheta}$.
For \texttt{Direct-Map} and \texttt{Perceptual-Map}, as in~\cite{zamir2020robust} we use the whole SegNet as the cross-task mapping functions. 
We use the same data augmentations from the updated implementation in~\cite{liu2019end}\footnote{https://github.com/lorenmt/mtan}, \ie random crops and rand horizontal flips.

\begin{table}[h]
	\centering
	
    \resizebox{1\textwidth}{!}
    {
		\begin{tabular}{lccccccccc}

		    \toprule
		    Dataset & \# Train & \# Test & Segmentation & Depth & Human Parts & Normals & Saliency & Edges \\
		    \midrule
		    Cityscapes~\cite{cordts2016cityscapes} & 2975 & 500 & \Checkmark & \Checkmark & -\ - & -\ - & -\ - & -\ -\\
		    NYU-v2~\cite{silberman2012indoor} & 795 & 654 & \Checkmark & \Checkmark & -\ - & \Checkmark & -\ - & -\ - \\
		    PASCAL~\cite{chen2014detect} & 4998 & 5105 & \Checkmark & -\ - & \Checkmark & \Checkmark & \Checkmark & \Checkmark \\
			\bottomrule
		\end{tabular}%
			}
		\vspace{-0.25cm}
		\caption{Details of multi-task benchmarks.}
		\label{supptab:datasets}
\end{table}%

\begin{table}[t]
	\centering
	
    \resizebox{1\textwidth}{!}
    {
		\begin{tabular}{lccccccccc}

		    \toprule
		    \multirow{2}{*}{Dataset} & \multirow{2}{*}{\# label} & \multicolumn{6}{c}{\# labelled images} \\
		    % \cmidrule{3-8}
		    & & Segmentation & Depth & Human Parts & Normals & Saliency & Edges \\
		    \midrule
		    Cityscapes~\cite{cordts2016cityscapes} & one & 1487 & 1488 & -\ - & -\ - & -\ - & -\ - \\
		    \midrule
		    \multirow{2}{*}{NYU-v2~\cite{silberman2012indoor}} & random & 392 & 408 & -\ - & 385 & -\ - & -\ - \\
		    & one & 265 & 265 & -\ -  & 265 & -\ -  & -\ - \\
		    \midrule
		    \multirow{2}{*}{PASCAL~\cite{chen2014detect}} & random & 2450 & -\ - & 2553 & 2480 & 2445 & 2557 \\
		    & one & 1000 & -\ - & 999 & 1000 & 1000 & 999 \\
			\bottomrule
		\end{tabular}%
			}
		\vspace{-0.25cm}
		\caption{Details about multi-task partially supervised learning settings in three benchmarks used in this work. `random' means the random-label setting where each training image has a random number of task labels and `one' indicates the one-label setting where each training image is annotated with one task label. `\# labelled images' shows the number of images containing labels for each task, \eg segmentation.}
		\label{supptab:psl}
\end{table}%
% \vspace{-0.2cm}

\paragraph{NYU-v2.} The dataset~\cite{silberman2012indoor} contains 795 training images and 654 test images. To evaluate the multi-task partially supervised learning, we consider one-label and random-label settings. For one-label setting, we randomly select and keep label for only 1 task for each training image, resulting in 265 images with annotation for segmentation, 265 images labelled for depth estimation and 265 images for surface normal. For random-label setting, we randomly select and keep labels for at least 1 and at most 2 tasks (1.49 labels per image), \ie 392 images for semantic segmentation, 408 images for depth estimation, 385 images for surface normal, as shown in \cref{supptab:psl}.

We follow the training and evaluation protocol in~\cite{liu2019end} and we use the the SegNet~\cite{badrinarayanan2017segnet} as the MTL backbone for all methods. 
As in ~\cite{liu2019end}, we use cross-entropy loss for semantic segmentation, l1-norm loss for depth estimation and cosine similarity loss for surface normal estimation, use the same optimizer and hyper-parameters, \ie Adam optimizer with a learning rate of 0.0001. 
We train the all model for 200 epochs with a batch size of 2 and halve the learning rate at the 100-th epoch and employ the same evaluation metrics, mean intersection over union (mIoU), absolute error (aErr) and mean error (mErr) in the predicted angles to evaluate the semantic segmentation, depth estimation and surface normals estimation task, respectively as in~\cite{liu2019end}. 

We use the encoder of SegNet architecture for the joint pairwise task mapping ($\bar{m}_{\vartheta}$) and one convolutional layer as task specific input layer in $\bar{m}_{\vartheta}$. 
For \texttt{Direct-Map} and \texttt{Perceptual-Map}, as in~\cite{zamir2020robust} we use the whole SegNet as the cross-task mapping functions. To regularize training, we use the exact same data augmentations from the updated implementation from~\cite{liu2019end}, \eg random crops and rand horizontal flips augmentations.

% \paragraph{Implementation details on PASCAL.} 
\paragraph{PASCAL-context.} The dataset~\cite{chen2014detect} contains 4998 training images and 5105 testing images for five tasks, \ie semantic segmentation, human parts segmentation, surface normal, saliency detection and edge detection. We consider two partially supervised learning settings, random-label and one-label setting. For one-label setting, we have 1 label per image, \ie 1000, 999, 1000, 1000, 999 labelled images for semantic segmentation, human parts, surface normal, saliency and edge detection, respectively. In random-label setting, we randomly sample and keep labels for at least 1 and at most 4 tasks (2.50 labels per image), resulting in 2450, 2553, 2480, 2445, 2557 labelled images for semantic segmentation, human parts, surface normal, saliency and edge detection, respectively, as shown in \cref{supptab:psl}.

We follow exactly the  same training, evaluation protocol and implementation in~\cite{vandenhende2021multi} and employ the ResNet-18~\cite{he2016deep} as the encoder shared across all tasks and Atrous Spatial Pyramid Pooling (ASPP)~\cite{chen2018encoder} module as task-specific heads. 
We use the same hyper-parameters, \eg learning rate, augmentation, loss functions, loss weights in \cite{vandenhende2021multi}. More specifically, we use Adam as the optimizer with a learning rate of 0.0001 and a weight decay of 0.0001. As in~\cite{vandenhende2021multi} all experiments are performed using pre-trained ImageNet weights. We train all multi-task learning methods for 100 epochs with a batch size of 6 and we anneal the learning rate using the `poly' learning rate scheduler as in~\cite{vandenhende2021multi,chen2017deeplab}. We follow~\cite{vandenhende2021multi} and use fixed loss weights for training all multi-task learning methods, \ie the loss weight is 1, 2, 10, 5, 50 for semantic segmentation, human parts segmentation, surface normal estimation, saliency detection and edge detection, respectively. Please refer to~\cite{vandenhende2021multi} for more details.
For evaluation metrics, we use the optimal dataset F-measure (odsF)~\cite{martin2004learning} for edge detection, the standard mean intersection over union (mIoU) for semantic segmentation, human part segmentation and saliency estimation are evaluated,  mean error (mErr) for surface normals.
We modify the ResNet-18 to have task specific input layers (one convolutional layer for each task) before the residual blocks as the mapping function $\bar{m}_{\vartheta}$ in our method.

\paragraph{Multi-task performance.}
Following prior work~\cite{vandenhende2021multi}, we also report the multi-task performance $\bigtriangleup$MTL of the multi-task learning model as the average per-task drop in performance w.r.t. the single-task baseline:
\begin{equation}
\bigtriangleup\text{MTL}=\frac{1}{K}\sum_{t=1}^{K}(-1)^{\ell_i}(P^{mtl}_{t}-P^{stl}_{t})/P^{stl}_{t},
\end{equation} where $\ell_i=1$ if a lower value of $P_{t}$ means better performance for metric of task $t$, and 0 otherwise.

\section{More results}
Here, we report more results from single-task learning (STL) model, Contrastive-Loss and Discriminator-Loss and also qualitative results.

\subsection{Quantitative results}

\paragraph{Results on Cityscapes.}
Here, we report the results on Cityscapes for only \emph{one} label setting as there are two tasks in total in \cref{supptab:citys}. We also report results of single-task learning models which are used to compute the multi-task performance ($\bigtriangleup$MTL) to better analyze the results as in~\cite{vandenhende2021multi}. The performance of MTL methods are worse than single-task learning models for some tasks as the MTL models have less capacity and there is a problem of imbalanced optimization etc as discussed in~\cite{kendall2018multi,li2020knowledge,vandenhende2021multi}.

The results of MTL model learned with SL when all task labels are available for training to serve as a strong baseline for multi-task learning methods.
In the partial label setting (one task label per image), the performance of the SL baseline drops substantially compared to its performance in full supervision setting. 
While the SSL baseline, by extracting task-specific information from unlabelled tasks, improves over SL, further improvements are obtained by exploiting cross-task consistency in various ways except \DL.
The methods learn mappings from one task to another one (\PM\ and \DM) surprisingly perform better than the ones learning joint space mapping functions (\CL\ and \DL), possibly due to insufficient number of negative samples.
Finally, the best results (\eg the best multi-task performance $\bigtriangleup \text{MTL}$) are obtained with our method that can exploit cross-task relations more efficiently through joint pairwise task mappings with the proposed regularization.
Interestingly, our method also outperforms the SL baseline that has access to all the task labels, showing the potential information in the cross-task relations. 

\begin{table}[h]
	\centering
	
    \resizebox{1\textwidth}{!}
    {
		\begin{tabular}{cllccccc}

		    \toprule
		    \# label & Type & Method & Seg. (IoU) $\uparrow$ & Depth (aErr) $\downarrow$ & $\bigtriangleup$MTL $\uparrow$ \\
		    \midrule
		    \multirow{2}{*}{full} & STL & Supervised Learning & 74.19 & 0.0124 & +0.00 \\
		    & MTL & Supervised Learning & 73.36 & 0.0165 & -17.00 \\
		    \midrule
		    \multirow{8}{*}{one} & STL & Supervised Learning & 70.26 & 0.0141  & +0.00 \\
		    \cmidrule{2-6}
		    & \multirow{7}{*}{MTL} & Supervised Learning & 69.50 & 0.0186 & -16.55 \\
		    & & Semi-supervised Learning & 71.67 & 0.0178 & -12.22 \\
		    & & Perceptual-Map & 72.82 & 0.0169 & -8.37 \\
		    & & Direct-Map & 72.33 & 0.0179 & -11.94 \\
		    & & Contrastive-Loss & 71.79 & 0.0183 & -13.77 \\
		    & & Discriminator-Loss & 68.94 & 0.0208 & -24.95 \\
			\cmidrule{3-6}	
		    & & Ours & {\bf 74.90} & {\bf 0.0161} & {\bf -3.81} \\
			\bottomrule
		\end{tabular}%
			}
		\vspace{-0.25cm}
		\caption{Multi-task learning results on Cityscapes. `one' indicates each image is randomly annotated with one task label. `STL' means single task learning and `MTL' indicates multi-task learning.}
		\label{supptab:citys}
\end{table}%

\paragraph{Results on Cityscapes with larger images.}
We also provide results for $256\times 512$ setting in \cref{supptab:largecitys}. Performance of all methods improve significantly compared to their ones using small images (in~\cref{supptab:citys}) and our method achieves significant improvement over the baselines. 

\begin{table}[h]
	\centering
	
    \resizebox{1\textwidth}{!}
    {
		\begin{tabular}{cllccccc}

		    \toprule
		    \# label & Type & Method & Seg. (IoU) $\uparrow$ & Depth (aErr) $\downarrow$ & $\bigtriangleup$MTL $\uparrow$ \\
		    \midrule
		    \multirow{4}{*}{one} & STL & Supervised Learning & 77.97 & 0.0126 & +0.00 \\
		    \cmidrule{2-6}
		    & \multirow{3}{*}{MTL} & Supervised Learning & 77.71 & 0.0165 & -15.95  \\
		    & & Semi-supervised Learning & 79.24 & 0.0161 & -13.38 \\
			\cmidrule{3-6}	
		    & & Ours & {\bf 82.41} & {\bf 0.0143} & {\bf -4.08} \\
			\bottomrule
		\end{tabular}%
			}
		\vspace{-0.25cm}
		\caption{\footnotesize Multi-task learning results on Cityscapes using $256\times 512$ images. `one' indicates each image is randomly annotated with one task label. `STL' means single task learning and `MTL' indicates multi-task learning.}
		\label{supptab:largecitys}
\end{table}%

\paragraph{Results on NYU-v2}
Here, we evaluate our method and related methods in the \emph{random} and \emph{one} label settings on NYU-v2 and we report the results in \cref{supptab:nyuv2}. We also report results of single-task learning models which are used to compute the multi-task performance ($\bigtriangleup$MTL) to better analyze the results as in~\cite{vandenhende2021multi}.

While we observe a similar trend across different methods, overall the performances are lower in this benchmark possibly due to fewer training images than CityScapes.
As expected, the performance in random-label setting is better than the one in one-label setting, as there are more labels available in the former.
While the best results are obtained with SL trained on the full supervision, our method obtains the best performance (\eg best results on all tasks and the best multi-task performance) among the partially supervised methods.
Here SSL improves over SL trained on the partial labels and cross-task consistency is beneficial except for \DM\ in the one label setting and Discriminator-Loss, possibly because the dataset is too small to learn accurate mappings between two tasks, while our method is more data-efficient and more successful to exploit the cross-task relations. In random-label setting, where images might have labels for more than one task, we also report our method also leveraging the labelled corss-task relations (`Ours+' ) in \cref{supptab:nyuv2} and it can indeed further boost the average performance.

\begin{table}[h]
	\centering
	
    \resizebox{1\textwidth}{!}
    {
		\begin{tabular}{clcccccccccc}

		    \toprule
		     \# labels & Type & Method & Seg. (IoU) $\uparrow$ & Depth (aErr) $\downarrow$ & Norm. (mErr) $\downarrow$ & $\bigtriangleup$MTL $\uparrow$ \\
		    \midrule
		    \multirow{2}{*}{full} & STL & Supervised learning & 37.45 & 0.6079 & 25.94 & +0.00 \\
		    & MTL & Supervised learning & 36.95 & 0.5510 & 29.51 & -1.92 \\
		    \midrule
		    \multirow{10}{*}{random} & STL & Supervised Learning & 28.72 & 0.7540 & 28.95 & +0.00  \\
		    \cmidrule{2-7}
		    & \multirow{7}{*}{MTL} & Supervised Learning & 27.05 & 0.6624 & 33.58 & -3.23 \\
		    & & Semi-supervised Learning & 29.50 & 0.6224 & 33.31 & +1.70 \\
		    & & Perceptual-Map & 32.20 & 0.6037 & 32.07 & +7.10 \\
		    & & Direct-Map & 29.17 & 0.6128 & 33.63 & +1.38 \\
		    & & Contrastive-Loss & 30.75 & 0.6143 & 32.05 & +4.96 \\
		    & & Discriminator-Loss & 26.76 & 0.6354 & 33.13 & -1.84 \\
		    \cmidrule{3-7}
		    & & Ours & 34.26 & 0.5787 & {\bf 31.06} & +11.81 \\
		    & & Ours+ & {\bf 34.91} & {\bf 0.5738} & 31.20 & {\bf +12.57} \\
		    \midrule
		    \multirow{9}{*}{one} & STL & Supervised Learning & 24.71 & 0.7666 & 30.14 & +0.00 \\
		    \cmidrule{2-7}
		    & \multirow{7}{*}{MTL} & Supervised Learning & 25.75 & 0.6511 & 33.73 & +1.14 \\
		    & & Semi-supervised Learning & 27.52 & 0.6499 & 33.58 & +3.16 \\
		    & & Perceptual-Map & 26.94 & 0.6342 & 34.30 & +2.31 \\
		    & & Direct-Map & 19.98 & 0.6960 & 37.56 & -12.86 \\
		    & & Contrastive-Loss & 26.65 & 0.6387 & 34.69 & +1.31 \\
		    & & Discriminator-Loss & 25.68 & 0.6566 & 34.02 & +0.04 \\
		    \cmidrule{3-7}
		    & & Ours & {\bf 30.36} & {\bf 0.6088} & {\bf 32.08} & {\bf +10.24} \\
			\bottomrule
		\end{tabular}%
			}
		\vspace{-0.25cm}
		\caption{Multi-task learning results on NYU-v2. `random' indicates each image is annotated with a random number of task labels and `one' means each image is randomly annotated with one task. `STL' means single task learning and `MTL' indicates multi-task learning.}
		\label{supptab:nyuv2}
\end{table}%
% \vspace{0.1cm}

\paragraph{Results on PASCAL.}
We evaluate all methods on PASCAL-Context, in both label settings, which contains wider variety of tasks than the previous benchmarks and report the results in \cref{supptab:pascal}. As in Cityscapes and NYU-v2, we also report results of single-task learning models which are used to compute the multi-task performance ($\bigtriangleup$MTL) to better analyze the results as in~\cite{vandenhende2021multi}. 

As the required number of pairwise mappings for \DM\ and \PM\ grows quadratically (20 mappings for 5 tasks), we omit these two due to their high computational cost and compare our method only to SL, SSL, Contrastive-Loss and Discriminator-Loss baselines.
We see that the SSL baseline improves the performance over SL in random-label setting, however, it performs worse than the SL in one label setting, when there are 60\% less labels.
By leveraging cross-task consistency, Contrastive-Loss and Discriminator-Loss obtains better performance than the SL baseline in one label setting while they get similar multi-task performance to the SL baseline in random label setting.
Again, by exploiting task relations, our method obtains better or comparable results to second best method, \ie SSL, while the gains achieved over SL and SSL are more significant in the low label regime (one-label).
Interestingly, SSL and our method obtain comparable results in random-label setting which suggests that relations across tasks are less informative than the ones in CityScape and NYUv2.

\begin{table}[h]
	\centering
	
% 	\resizebox{15cm}{}
    \resizebox{1.0\textwidth}{!}
    {
		\begin{tabular}{cllccccccccc}

		    \toprule
		    \# labels & Type & Method & Seg. (IoU) $\uparrow$ & H. Parts (IoU) $\uparrow$ & Norm. (mErr) $\downarrow$ & Sal. (IoU) $\uparrow$ & Edge (odsF) $\uparrow$ & $\bigtriangleup$MTL $\uparrow$ \\
		    \midrule
		    \multirow{2}{*}{full} & STL & Supervised Learning & 66.4 & 58.9 & 13.9 & 66.7 & 68.3 & +0.00 \\
		    & MTL & Supervised Learning & 63.9 & 58.9 & 15.1 & 65.4 & 69.4 & -2.75 \\
		    \midrule
		    \multirow{7}{*}{random} & STL & Supervised Learning & 60.9 & 55.3 & 14.7 & 64.8 & 66.8 & +0.00 \\
		    \cmidrule{2-9}
		    & \multirow{5}{*}{MTL} & Supervised Learning & 58.4 & 55.3 & 16.0 & 63.9 & {\bf 67.8} & -2.67 \\
		    & & Semi-supervised Learning & {\bf 59.0} & {\bf 55.8} & {\bf 15.9} & {\bf 64.0} & 66.9 & -2.44 \\
		    & & Contrastive-Loss & {\bf 59.0} & 55.3 & 16.0 & 63.8 & {\bf 67.8} & -2.44 \\
		    & & Discriminator-Loss & 57.9 & 55.2 & 16.2 & 63.4 & 67.4 & -3.35 \\
		    \cmidrule{3-9}
		    & & Ours & {\bf 59.0} & 55.6 & {\bf 15.9} & {\bf 64.0} & {\bf 67.8} & {\bf -2.15} \\
		    \midrule
		    \multirow{7}{*}{one} & STL & Supervised Learning & 47.7 & 56.2 & 16.0 & 61.9 & 64.0 & +0.00 \\
		    \cmidrule{2-9}
		    & \multirow{5}{*}{MTL} & Supervised Learning & 48.0 & 55.6 & 17.2 & 61.5 & 64.6 & -1.34 \\
		    & & Semi-supervised Learning & 45.0 & 54.0 & {\bf 16.9} & {\bf 61.7} & 62.4 & -3.02 \\
		    & & Contrastive-Loss & 48.5 & 55.4 & 17.1 & 61.3 & 64.6 & -1.25 \\
		    & & Discriminator-Loss & 48.2 & {\bf 56.0} & 17.1 &  61.7 & 64.7 & -1.04 \\
		    \cmidrule{3-9}
		    & & Ours & {\bf 49.5} & 55.8 & 17.0 & {\bf 61.7} & {\bf 65.1} & {\bf -0.40} \\
			\bottomrule
		\end{tabular}%
			}
		\vspace{-0.25cm}
		\caption{Multi-task learning results on PASCAL. `random' indicates each image is annotated with a random number of task labels and `one' means each image is randomly annotated with one task. `STL' means single task learning and `MTL' indicates multi-task learning.}
		\label{supptab:pascal}
\end{table}%
% \vspace{0.1cm}

\paragraph{Learning from partial and imbalanced task labels.}

We also evaluate our method and baselines in an imbalanced partially supervised setting in Cityscapes, where we assume the ratio of labels for each task are imbalanced, \eg we randomly sample 90\% of images to be labeled for semantic segmentation and only 10\% images having labels for depth and we denote this setting by the label ratio between segmentation and depth (Seg.:Depth = 9:1). 
The opposite case (Seg.:Depth = 1:9) is also considered. We report the results in \cref{supptab:citysimbalance}, where we also report results of single-task learning models which are used to compute the multi-task performance ($\bigtriangleup$MTL) to better analyze the results as in~\cite{vandenhende2021multi}. 

\begin{table}[h]
	\centering
	
% 	\resizebox{15cm}{}
    \resizebox{1\textwidth}{!}
    {
		\begin{tabular}{cllccccc}

		    \toprule
		    \#labels & Type & Method & Seg. (IoU) $\uparrow$ & Depth (aErr) $\downarrow$ & $\bigtriangleup$MTL $\uparrow$ \\
		    \midrule
		    \multirow{2}{*}{full} & STL & Supervised learning & 74.19 & 0.0124 & +0.00 \\
		    & MTL & Supervised Learning & 73.36 & 0.0165 & -17.00 \\
		    \midrule
		    \multirow{9}{*}{1:9} & STL & Supervised learning & 62.23 & 0.0126 & +0.00 \\
		    \cmidrule{2-6}
		    & \multirow{7}{*}{MTL} & Supervised Learning & 63.37 & 0.0161 & -13.07 \\
		    & & Semi-supervised Learning & 64.40 & 0.0179 & -19.36 \\
		    & & Perceptual-Map & 68.84 & 0.0141 & -0.68 \\
		    & & Direct-Map & 67.04 & 0.0153 & -6.90 \\
		    & & Contrastive-Loss & 67.12 & 0.0151 & -5.95 \\
		    & & Discriminator-Loss & 68.92 & 0.0144 & -1.80 \\
		    \cmidrule{3-6}
		    & & Ours & {\bf 71.89} & {\bf 0.0131} & {\bf +5.63} \\
		    \midrule
		    \multirow{9}{*}{9:1} & STL & Supervised learning & 72.62 & 0.0191 & +0.00 \\
		    \cmidrule{2-6}
		    & \multirow{7}{*}{MTL} & Supervised learning & 72.77 & 0.0250 & -15.25 \\
		    & & Semi-supervised Learning & 72.97 & 0.0395 & -53.11 \\
		    & & Perceptual-Map & 73.36 & 0.0237 & -11.34 \\
		    & & Direct-Map & 73.13 & 0.0288 & -19.38 \\
		    & & Contrastive-Loss & 73.75 & 0.0243 & -12.86 \\
		    & & Discriminator-Loss & 72.97 & 0.0248 & -14.65\\
		    \cmidrule{3-6}
		    & & Ours & {\bf 74.23} & {\bf 0.0235} & {\bf -10.23} \\
			\bottomrule
		\end{tabular}%
			}
		\vspace{-0.25cm}
		\caption{Multi-task learning results on Cityscapes. `\#label' indicates the number ratio of labels for segmentation and depth, \eg `1:9' means we have 10\% of images annotated with segmentation labels and 90\% of images have depth groundtruth. `STL' means single task learning and `MTL' indicates multi-task learning.}
		\label{supptab:citysimbalance}
\end{table}%

The performance of supervised learning (SL) on the task with partial labels drops significantly. Though SSL improves the performance on segmentation, its performance on depth drops in both cases. Different from SSL, Direct-Map, Contrastive-Loss and Discriminator-Loss improves the performance on both tasks in 1:9 setting while their performance on depth drop in the 9:1 case.
In contrast to SL and the baselines, our method and Perceptual-Map obtain better results on all tasks in both settings by learning cross-task consistency while our method obtains the best performance (\ie best results in all tasks and best multi-task performance, $\bigtriangleup$MTL) by joint space mapping. 
This demonstrates that our model can successfully learn cross-task relations from unbalanced labels thanks to its task agnostic mapping function which can share parameters across multiple task pairs.

\paragraph{Cross-task consistency learning in conventional semi-supervised learning.}
We evaluate our method and SSL baseline on conventional SSL setting where $\frac{1}{3}$ of training data in NYU-v2 are labeled for all tasks and $\frac{2}{3}$ are unlabeled, and report the results in \cref{supptab:sslnyuv2}. In this setting, our method obtains better performance than SL and SSL. We will include a more detailed analysis in the final paper.

\begin{table}[h!]
	\centering
	
    \resizebox{1\textwidth}{!}
    {
		\begin{tabular}{clcccccccccc}

		    \toprule
		    Type & Method & Seg. (IoU) $\uparrow$ & Depth (aErr) $\downarrow$ & Norm. (mErr) $\downarrow$ & $\bigtriangleup$MTL $\uparrow$ \\
		    \midrule
		    \multirow{3}{*}{MTL} & Supervised Learning & 24.78 & 0.6681 & 33.90 & +1.48 \\
		    & Semi-supervised Learning & 26.09 & 0.6510 & 33.60 & +4.37 \\
		    \cmidrule{2-6}
		    & Ours & {\bf 28.43} & {\bf 0.6366} & {\bf 33.01} & {\bf +8.83} \\
			\bottomrule
		\end{tabular}%
			}
		\vspace{-0.25cm}
		\caption{\footnotesize Multi-task learning results on NYU-v2 in SSL setting where $\frac{1}{3}$ of training data in NYU-v2 are labeled for all tasks and $\frac{2}{3}$ are unlabeled. `MTL' indicates multi-task learning.}
		\label{supptab:sslnyuv2}
\end{table}%

\paragraph{Cross-task consistency learning with full supervision.}
Our method can also be applied to fully-supervised learning setting where all task labels are available for each sample by mapping one task's prediction and another task's ground-truth to the joint space and measuring cross-task consistency in the joint space. We applied our method to NYU-v2 and compare it with the single task learning (STL) networks, vanilla MTL baseline, recent multi-task learning methods, \ie MTAN~\cite{liu2019end}, X-task~\cite{zamir2020robust}, and several methods focusing on loss weighting strategies, \ie Uncertainty~\cite{kendall2018multi}, GradNorm~\cite{chen2018gradnorm}, MGDA~\cite{sener2018multi} and DWA~\cite{liu2019end} in \cref{supptab:nyuv2full}. Here, we also report the multi-task performance ($\bigtriangleup$MTL) of all MTL methods.

\begin{table}[h!]
	\centering
	
% 	\resizebox{15cm}{}
    \resizebox{1\textwidth}{!}
    {
		\begin{tabular}{lccccccccccc}

		    \toprule
		    Method & Seg. (IoU) $\uparrow$ & Depth (aErr) $\downarrow$ & Norm. (mErr) $\downarrow$ & $\bigtriangleup$MTL \\
		    \midrule
		    STL & 37.45 & 0.6079 & 25.94 & +0.00 \\
		    \midrule
		    MTL & 36.95 & 0.5510 & 29.51 & -1.92 \\
		    MTAN~\cite{liu2019end} & 39.39 & 0.5696 & 28.89 & +0.03 \\
		    X-task~\cite{zamir2020robust} & 38.91 &  0.5342 & 29.94 & +0.89 \\
		    Uncertainty~\cite{kendall2018multi} & 36.46 & 0.5376 & 27.58 & +0.86 \\
		    GradNorm~\cite{chen2018gradnorm} & 37.19 & 0.5775 & 28.51 & -1.86 \\
		    MGDA~\cite{sener2018multi} & 38.65 & 0.5572 & 28.89 & +0.06 \\
		    DWA~\cite{liu2019end} & 36.46 & 0.5429 & 29.45 & -1.82 \\
		    \midrule
		    Ours & 41.00 & 0.5148 & 28.58 & +4.88 \\
		    Ours + Uncertainty & {\bf 41.09} & {\bf 0.5090} & {\bf 26.78} & {\bf +7.57} \\
			\bottomrule
		\end{tabular}%
			}
		\vspace{-0.25cm}
		\caption{Multi-task fully-supervised learning results on NYU-v2. `STL' indicates standard single-task learning and `MTL' means the standard multi-task learning network.}
		\label{supptab:nyuv2full}
\end{table}%

MTL, MTAN, X-task and Ours are trained with uniform loss weights. 
We see that our method (Ours) performs better than the other methods with uniform loss weights, \eg MTAN and X-task, where X-task regularizes cross-task consistency by learning perceptual loss with pre-trained cross-task mapping functions. 
This shows that cross-task consistency is informative even in the fully supervised case and our method is more effective for learning cross-task consistency. 
Compared to recent loss weighting strategies, our method (Ours) obtains better multi-task performance ($\bigtriangleup$MTL) and better performance on segmentation and depth estimation than other methods while slightly worse on normal estimation compared with GradNorm and Uncertainty. 
This is because the loss weighting strategies enable a more balanced optimization of multi-task learning model than uniformly loss weighting. 
Thus when we incorporate the loss weighing strategy of Uncertainty~\cite{kendall2018multi} to our method, \ie (Ours + Uncertainty), our method obtains further improvement and outperforms both GradNorm and Uncertainty, \eg `Ours + Uncertainty' obtains the best multi-task performance (+7.57).

\subsection{Qualitative results}
\begin{figure}[h!]
\begin{center}
\includegraphics[width=1\linewidth]{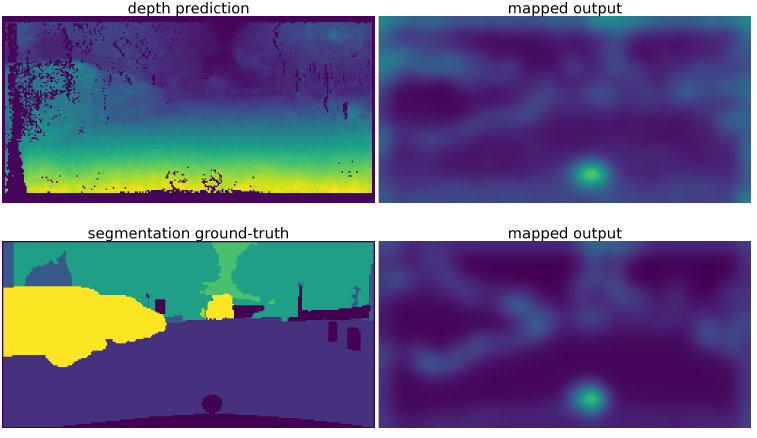}
\end{center}
\vspace{-0.65cm}
\caption{Intermediate feature map of the mapping function of the task-pair (segmentation to depth) of one example in Cityscapes. The first column shows the prediction or ground-truth and the second column present the corresponding mapped feature map (output of the mapping function's last second layer ).}
\label{suppfig:mappedcitys}
\end{figure}

\begin{figure}[h!]
\begin{center}
\includegraphics[width=1\linewidth]{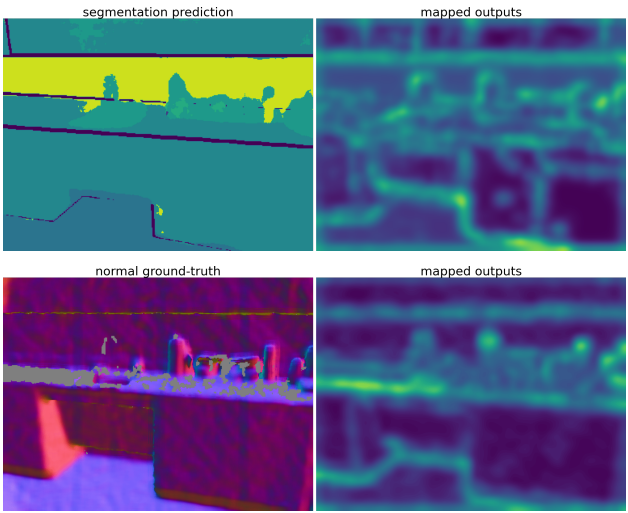}
\end{center}
\vspace{-0.65cm}
\caption{Intermediate feature map of the mapping function of the task-pair (segmentation to surface normal) of one example in NYU-v2. The first column shows the prediction or ground-truth and the second column present the corresponding mapped feature map (output of the mapping function's last second layer ).}
\label{suppfig:mappednyu}
\end{figure}

% \vspace{-0.1cm}
Here, we present some qualitative results.
\paragraph{Mapped outputs.}
Here, we visualize the intermediate feature maps of $m^{s\rightarrow st}$ and $m^{t\rightarrow st}$ for one example in Cityscapes in \cref{suppfig:mappedcitys} where $s$ and $t$ correspond to segmentation and depth estimation respectively and one example in NYU-v2 in \cref{suppfig:mappednyu} where $s$ and $t$ correspond to segmentation and surface normal estimation respectively.
We observe that the functions map both task labels to a joint pairwise space where the common information is around object boundaries, which in turn enables the model to produce more accurate predictions for both tasks.

% \vspace{-0.2cm}

\begin{figure}[h!]
\begin{center}
\includegraphics[width=1.0\linewidth]{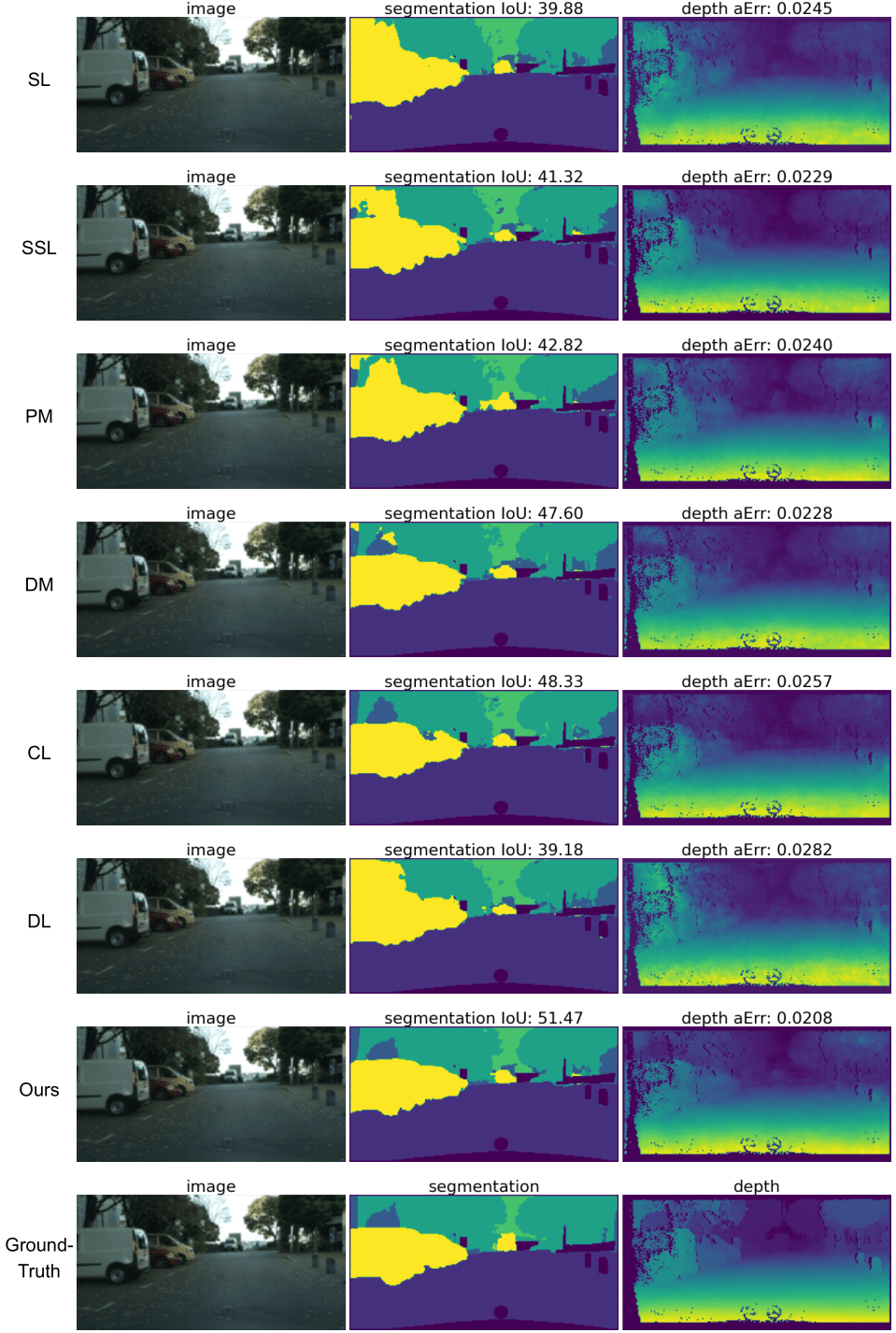}
\end{center}
\vspace{-0.65cm}
\caption{\textbf{Qualitative results on Cityscapes.} The fist column shows the RGB image, the second column plots the ground-truth or predictions with the IoU ($\uparrow$) score of all methods for semantic segmentation and we show the ground-truth or predictions with the absolute error ($\downarrow$) in the last column.}
\label{suppfig:predscitys}
\end{figure}

\begin{figure}
\begin{center}
\includegraphics[width=1.0\linewidth]{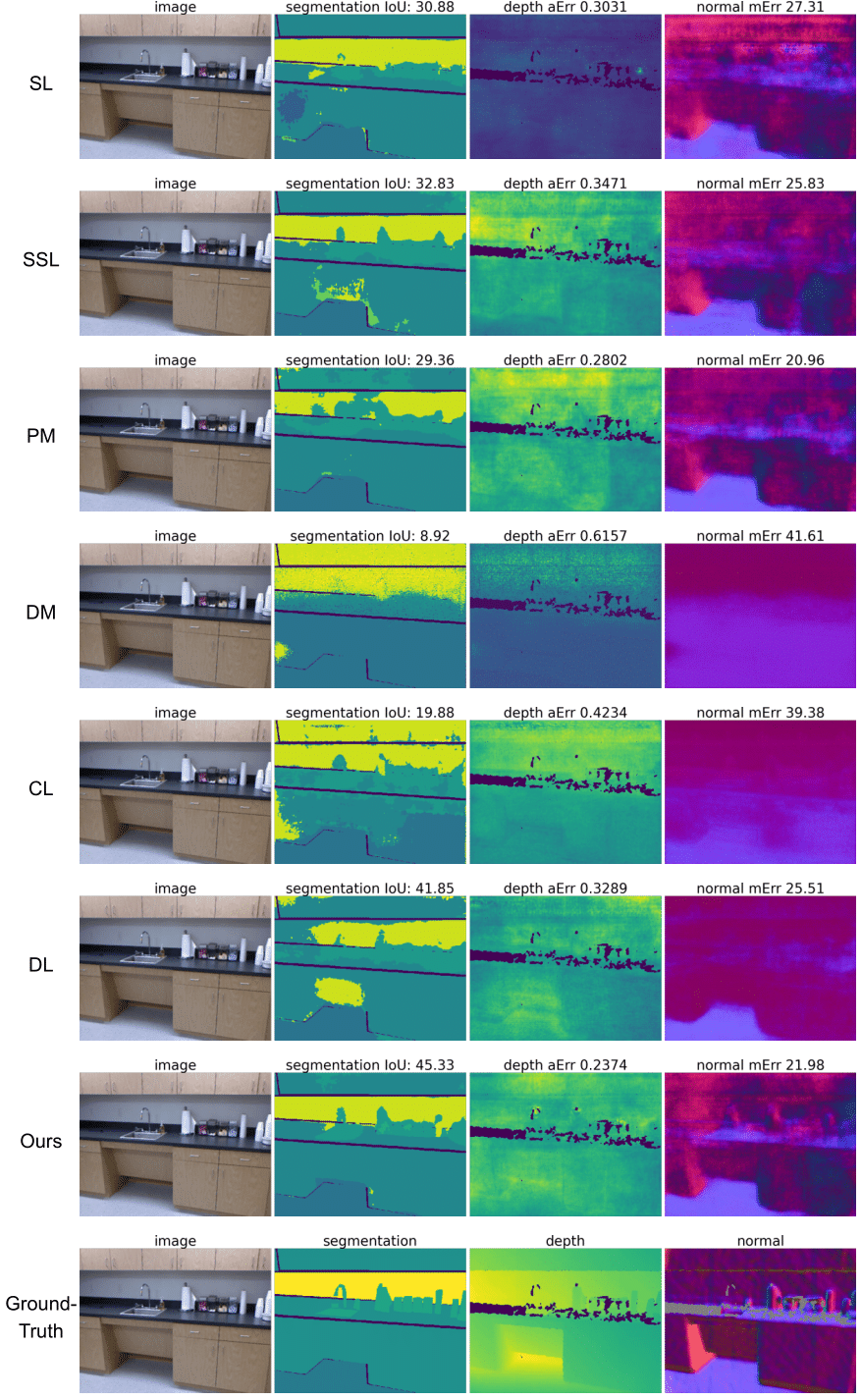}
\end{center}
\vspace{-0.65cm}
\caption{\textbf{Qualitative results on NYU-v2.} The fist column shows the RGB image, the second column plots the ground-truth or predictions with the IoU ($\uparrow$) score of all methods for semantic segmentation, the third column presents the ground-truth or predictions with the absolute error ($\downarrow$), and we show the prediction of surface normal with mean error ($\downarrow$) in the last column.}
\label{suppfig:predsnyu}
\end{figure}

\paragraph{Predictions.}
Finally we show qualitative comparisons between our method, SL and SSL baselines, Perceptual-Map (PM), Direct-Map (DM), Contrastive-Loss (CL) and Discriminator-Loss (DL) on Cityscapes in \cref{suppfig:predscitys} and on NYU-v2 in \cref{suppfig:predsnyu}. 
We can see that our method produces more accurate predictions by leveraging cross-task consistency. Specifically, in \cref{suppfig:predscitys}, compared with methods that do not leverage cross-task consistency, the prediction of segmentation and depth are improved by our method (top left region) and our results are more accurate than related baselines (PM, DM, CL and DL). In \cref{suppfig:predsnyu}, we can see that  SSL produces more accurate predictions on segmentation and surface normal than SL. And PM obtains more accurate results on depth and surface normal than SL. While they do not achieve consistent improvement on all three tasks, our method can improve the results consistently on three tasks which shows that our method is more effective on learning cross-task consistency for MTL from partially annotated data.

\end{document}